%% file: _main.tex
\input{_constants}
\arxiv 

\pdfoutput=1
\PassOptionsToPackage{svgnames}{xcolor}
\documentclass[10pt,twocolumn,letterpaper]{article}
\input{cvpr_header}

\definecolor{LightBlue}{HTML}{cce0ff}
\definecolor{Blue}{HTML}{99ccff}
\definecolor{DarkBlue}{HTML}{668cff}
\definecolor{LightTeal}{HTML}{B3FFFF}
\definecolor{MediumTeal}{HTML}{66FFFF}
\definecolor{DarkTeal}{HTML}{33CCCC}
\definecolor{LightYellow}{HTML}{FFFFCC}
\definecolor{MediumYellow}{HTML}{FFFF99}
\definecolor{DarkYellow}{HTML}{FFFF66}
\definecolor{lightgray}{gray}{0.9}

\begin{document}
\title{\paperTitle}
\author{\authorBlock}
\maketitle

\input{00_abstract}

\input{01_intro}
\input{02_related}

\input{problem_formulation}

\input{03_motivation}
\input{04_methods}
\input{05_experiment}

\input{10_conclusion}

{\small
\bibliographystyle{unsrtnat}
\bibliography{11_references}
}

\clearpage 
\appendix \input{12_appendix}

\end{document}

%% file: _constants.tex
\def\paperID{}
\def\confName{ICCV}
\def\confYear{2025}

\def\paperTitle{Sparrow: Data-Efficient Video-LLM with Text-to-Image Augmentation}

\def\authorBlock{
   Shukang Yin$^{1\dagger}$, ~Chaoyou Fu$^{2\dagger \ddagger}$\thanks{Project leader.}\hspace{1.1mm}, ~Sirui Zhao$^{1\ddagger}$\thanks{Equal contribution.}\hspace{1.0mm}, ~Chunjiang Ge$^{3}$, 
   ~Yan Yang$^{2}$, ~Yuhan Dai$^{1}$, \\Yongdong Luo$^{4}$, ~Tong Xu$^{1}$, ~Caifeng Shan$^{2}$, ~Enhong Chen$^{1}$\thanks{Corresponding authors.}\hspace{1.0mm}\\
$^1$USTC, $^2$NJU, $^3$THU, $^4$XMU
}

\newif\ifreview 
\newif\ifarxiv \newcommand{\arxiv}{\arxivtrue}
\newif\ifcamera 
\newif\ifrebuttal 

%% file: cvpr_header.tex
\ifreview \usepackage[review]{cvpr} \fi
\ifarxiv \usepackage[pagenumbers]{cvpr} \fi
\ifrebuttal \usepackage[rebuttal]{cvpr} \fi
\ifcamera \usepackage{cvpr} \fi

\input{_macros}  

\usepackage{xr-hyper}

\makeatletter
\newcommand*{\addFileDependency}[1]{
  \typeout{(#1)}
  \@addtofilelist{#1}
  \IfFileExists{#1}{}{\typeout{No file #1.}}
}

\makeatother

\definecolor{cvprblue}{rgb}{0.21,0.49,0.74}
\usepackage[pagebackref,breaklinks,colorlinks,allcolors=cvprblue]{hyperref}
\usepackage[capitalize]{cleveref}
\crefname{section}{Sec.}{Secs.}
\crefname{table}{Table}{Tables}
\crefname{figure}{Fig.}{Figs.}

\ifarxiv \crefname{appendix}{App.}{Apps.}
\else \crefname{appendix}{Suppl.}{Suppls.} \fi

%% file: _macros.tex
\usepackage[utf8]{inputenc}
\usepackage{graphicx}
\usepackage[numbers,sort&compress]{natbib}
\usepackage{indentfirst}   
\usepackage{bbding, pifont}
\usepackage{bm}
\usepackage{mathtools}
\usepackage{amsmath}	
\usepackage{amssymb}	
\usepackage{booktabs}
\usepackage{times}
\usepackage{microtype}
\usepackage{epsfig}
\usepackage{xcolor}
\usepackage{caption}
\usepackage{transparent}
\usepackage{float}
\usepackage{placeins}
\usepackage{color, colortbl}
\usepackage{stfloats}
\usepackage{enumitem}
\usepackage{tabularx}

\usepackage[normalem]{ulem}
\useunder{\uline}{\ul}{}

\usepackage{xstring}
\usepackage{multirow}
\usepackage{makecell}
\usepackage{xspace}
\usepackage{url}
\usepackage{subcaption}
\usepackage{tcolorbox}
\usepackage[hang,flushmargin]{footmisc}

\ifcamera \usepackage[accsupp]{axessibility} \fi

\def\eg{\emph{e.g}\onedot} 
\def\ie{\emph{i.e}\onedot} 
 
\def\etc{\emph{etc}\onedot} 
\def\wrt{w.r.t\onedot}

% ...

\ifarxiv  \fi

\newcommand{\R}[1]{{%
    \textbf{%
        \ifstrequal{#1}{1}{\textcolor{red}{R#1}}{%
        \ifstrequal{#1}{2}{\textcolor{blue}{R#1}}{%
        \ifstrequal{#1}{3}{\textcolor{magenta}{R#1}}{%
        \ifstrequal{#1}{4}{\textcolor{teal}{R#1}}{%
                           \textcolor{cyan}{R#1}%
        }}}}%
    }%
}}

\usepackage{amsmath}
\usepackage[linesnumbered,ruled,vlined]{algorithm2e}
\DontPrintSemicolon

\definecolor{LightBlue}{HTML}{cce0ff}
\definecolor{Blue}{HTML}{99ccff}
\definecolor{DarkBlue}{HTML}{668cff}
\definecolor{LightTeal}{HTML}{B3FFFF}
\definecolor{MediumTeal}{HTML}{66FFFF}
\definecolor{DarkTeal}{HTML}{33CCCC}
\definecolor{LightYellow}{HTML}{FFFFCC}
\definecolor{MediumYellow}{HTML}{FFFF99}
\definecolor{DarkYellow}{HTML}{FFFF66}
\definecolor{lightgray}{gray}{0.9}
\definecolor{perception_color}{HTML}{add5d9}
\definecolor{cognition_color}{HTML}{fab970}
\definecolor{info_color}{HTML}{cbc8c0}
\definecolor{mark_green}{HTML}{70AD47}

\newcommand{\plusvalue}[1]{\rlap{\hspace{0.1em}\scriptsize\textcolor{DarkGreen}{(+#1)}}}

\SetKwComment{Comment}{\color{green!50!black}\# }{}

\SetKwProg{Function}{def}{:}{}

\SetKwProg{For}{for}{:}{}
\SetKwProg{If}{if}{:}{}
\newcommand{\VarSty}[1]{\textnormal{\ttfamily\color{blue!90!black}#1}\unskip}

%% file: 00_abstract.tex
\begin{abstract}
Recent years have seen the success of Multimodal Large Language Models (MLLMs) in the domain of vision understanding.
The success of these models can largely be attributed to the dominant scaling law, which states that larger parameter sizes and data volumes contribute to better performance.
Notably, data scaling has been primarily driven by automatic data pipelines, which focus on the self-instruction of LLMs.
The paradigm has been taken for granted for quite some time, but the study of the effectiveness of scaling with these data has been neglected for a long time.
In this context, this work revisits scaling with synthetic data and focuses on developing video-LLMs from a data-centric perspective.
Our primary study approach involves fine-tuning pre-trained image-LLMs with video data and examining learning efficiency through data scaling.
Results from our preliminary experiments reveal a low learning efficiency phenomenon when simply scaling up video data samples, which, through our probing, can be ascribed to a lack of instruction diversity.
Aiming at this issue, we propose a data augmentation method called Sparrow, which synthesizes video-like samples from pure text instruction data. Mixing these synthetic samples with the video data enables a more efficient training scheme.
Through comprehensive experiments, we demonstrate that our proposed method achieves performance comparable to or even superior to that of baselines trained with significantly more samples.
Meanwhile, we find that incorporating these synthetic samples can enhance the performance of long video understanding without requiring training on long video data.
The code and data examples are available \href{https://github.com/VITA-MLLM/Sparrow}{here}.
\end{abstract}

%% file: 01_intro.tex
\begin{figure}[!thbp]\centering
\setlength{\abovecaptionskip}{1mm}
\setlength{\belowcaptionskip}{-6mm}
	\includegraphics[width=0.9\columnwidth]{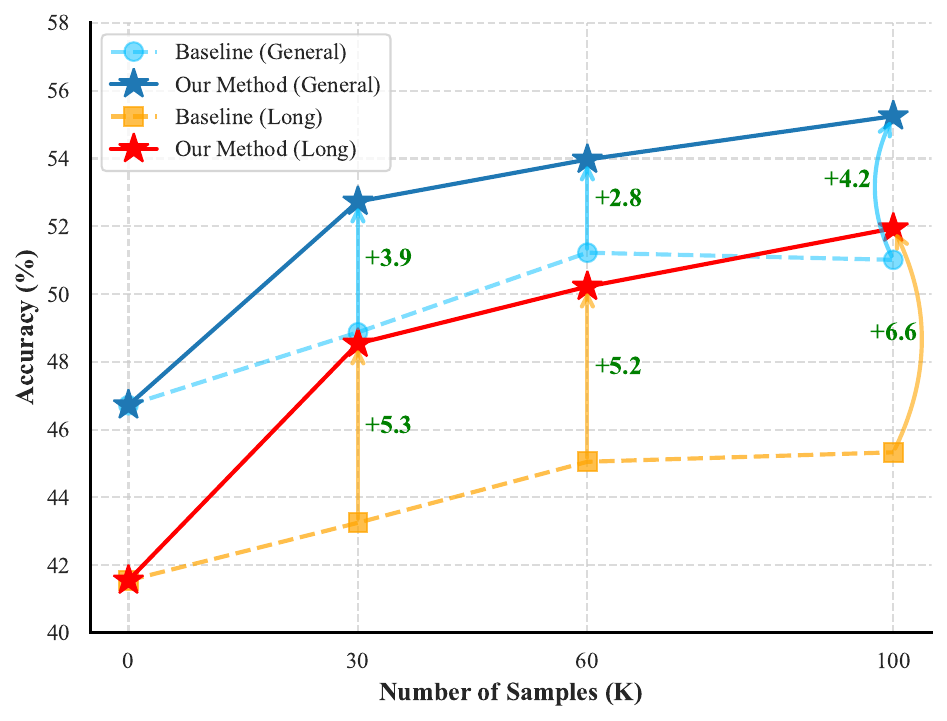}
	\caption{\textbf{Performance comparison with different schemes of data scaling.} We introduce \textbf{Sparrow}, a data-efficient training method for video-LLMs that achieves \textbf{``Train less, gain more''} in both general video and long video understanding performance. 0 sample indicates zero-shot inference with the image-LLM~\cite{minicpm}.}
	\label{fig:teaser}
\end{figure}

\section{Introduction}
\label{sec:intro}

The past few years have seen the rapid progress of Multimodal Large Language Models (MLLMs)~\cite{yin2024survey,xiong20253ur,wang2025multimodal,li2024lmeye,cao2025efficient,yin2025shapegpt,liu2024context,li2025etc}. Apart from solving traditional vision tasks (such as VQA), these models also excel in following user instructions and generalizing to new tasks. 
A mainstream paradigm for developing such models takes a two-stage training strategy. The first stage, pretraining, mainly serves to align vision modality with text and inject various kinds of visual knowledge into the model. In this stage, large-scale datasets of text-image pairs are often used, such as LAION~\cite{laion-5b} and CC~\cite{CC}, comprising a large proportion of the total compute and injecting abundant vision knowledge into models. Some methods also incorporate OCR and detection-related data to improve foundational capabilities~\cite{internvl, qwenvl}.
The second stage, instruction fine-tuning, adapts models to accommodate various tasks and helps generalize to new instructions. Training in this stage typically involves instruction data obtained from self-instruction or adaptation of task-specific datasets (\eg, VQA and chart understanding datasets).
Recently, researchers have shifted their focus from single-image models to more advanced ones that support video understanding. Borrowing successful experience from developing image models, some video counterparts are typically trained from scratch, following a similar two-stage training paradigm~\cite{videollama2,mvbench}. 
Apart from this path, some researchers utilize pre-trained image-LLMs instead. Typical approaches include zero-shot inference~\cite{ig-vlm,slowfast-llava,free-video-llm} and further fine-tuning~\cite{video-llava, mvbench, chatuniv, videollama2}.

Notably, the success of these models can be largely ascribed to the formidable scaling law, which puts emphasis on scaling up parameter size or data volume for better model performance.
For the data aspect, the scaling has mainly been driven by automatic data engines, which synthesize massive amounts of data without human labor.
Nevertheless, the characteristics of learning from these synthesized video data stand out as a critical yet underexplored topic.
Thus, in this work, we investigate the learning characteristics of video-LLM more deeply from a data scaling perspective.
Our preliminary data scaling experiments reveal a low data efficiency problem, that is, the performance gains from utilizing multiple times more data are marginal.
An inspection of data characteristics suggests this might be due to a lack of instruction diversity in the training corpus. 
To address this issue, we propose a data augmentation method, dubbed \textbf{Sparrow}\footnote{Inspiration taken from the swiftness of sparrows.}, to enrich the diversity of instruction. 
The basic idea is to synthesize video-like samples from textual data and mix these synthetic samples with the video samples. Specifically, we use existing text instruction data whose sample comprises a (long-context, instruction, answer) triplet. The long-context part is split into multiple segments and then further transformed into images, while the instruction and answer stay intact. Processed in this way, the synthetic samples have the same structure as video instruction data and can be incorporated seamlessly.

Comprehensive experiments demonstrate that our methods can facilitate data-efficient fine-tuning of image-LLMs for general video understanding and assist models in the comprehension of long videos. 
Specifically, using the same number of training samples, our method shows clear advantages over other data schemes. It even surpasses the baselines trained with many more samples, achieving high data efficiency (\Cref{fig:teaser}).
The contributions of this work include:
\begin{itemize}
    \item We investigate the fine-tuning approach for developing video-LLMs from a data perspective, and shed light on possible factors that lead to low learning efficiency.
    \item We propose a data augmentation method that improves the instruction diversity of training data and facilitates a more efficient training scheme.
    \item We perform comprehensive experiments to evaluate the proposed method and examine its key properties, paving the way for future research in this line.
\end{itemize}

%% file: 02_related.tex
\section{Related Work}
\label{sec:related}

\subsection{Multimodal Large Language Models}

\noindent \textbf{Image-LLMs.} 
To develop image-LLMs, the mainstream approach is to build upon powerful pre-trained LLMs and extend LLMs with the capability to perceive and reason with images~\cite{llava, qwenvl}. Based on a two-stage training recipe, \ie, image-text alignment training and instruction tuning, the developed model can fulfill a wide range of multimodal user queries and present its answers in user-friendly natural language sentences.

\noindent \textbf{Video-LLMs.} 
Following the success of image-LLMs, subsequent endeavors aim to expand the triumph to more intricate video understanding. 
Works like Video-ChatGPT~\cite{videochatgpt}, VTimeLLM~\cite{vtimellm}, PLLaVA~\cite{pllava}, and LLaVA-NeXT-Video~\cite{llava-next} attempt to further fine-tune image-LLMs to enhance video understanding capability. 
Other research~\cite{video-llava, mvbench, chatuniv, videollama2} explores training from pre-trained LLM, following the basic alignment-then-finetuning paradigm similar to image-LLM. These approaches usually involve joint training that mixes image and video data in the training corpus. 
In this study, we build upon pre-trained image-LLMs and enhance video understanding capabilities through fine-tuning.

\subsection{Long Video Understanding}
A fundamental challenge in long video comprehension lies in the effective modeling of long video frame sequences. To tackle this problem, two major technical directions have been actively investigated, namely context window extension and efficient video modeling.

\noindent \textbf{Context Window Extension.}
To accommodate a longer sequence of video frames with completeness, an intuitive approach is to extend the context window of the LLM backbone. Previous works have mainly adopted continued pretraining on the LLM before video training. 
For instance, LongVA~\cite{longva}, LongVILA~\cite{longvila}, and Kangaroo~\cite{kangaroo} perform continued training on long text corpora to accommodate more video tokens.

\noindent \textbf{Efficient Video Modeling.}
With the observation that visual information is often redundant in spatial and temporal dimensions, and that key frames are sparse in videos, another line of work explores efficient modeling of long videos. These works try to compress the video token sequences to fit in the original context window of language models.
Specifically, LLaMA-VID~\cite{llama-vid} proposes an attention-based module to compress each video frame into two tokens.
MovieChat~\cite{moviechat} designs a memory mechanism to cache video context and combines both short-term and long-term memory. 
TimeChat~\cite{timechat} combines image Q-Former and video Q-Former to compress video tokens within single frames and sliding windows of video, respectively.
Other works seek to retrieve frames most relevant to the user prompt and thus directly cut down the input length.
For example, KeyVideoLLM~\cite{keyvideollm} utilizes a pretrained CLIP encoder to calculate similarities between the query and video frames, and selects the most relevant frames based on the similarities.
Similarly, Video-RAG~\cite{videorag} further incorporates external tools to extract auxiliary texts from the video and augments the input with more multimodal context, such as OCR, ASR, and object detection information.

\subsection{Evaluation of Video Understanding}

Early methods~\cite{ig-vlm,videochatgpt,liang2025rethinking} are generally evaluated on traditional benchmarks like MSVD-QA~\cite{msvd-qa}, TGIF-QA~\cite{tgif-qa} and ActivityNet-QA~\cite{activity-net-qa}). These benchmarks are generally domain-specific and focus on certain basic skills, such as action recognition and repetition count, which lack comprehensiveness in both length coverage (especially in longer videos) and skill coverage. Moreover, the questions asked often involve shallow perception without deeper reasoning.

Recently, with the rise of benchmarks specifically designed for MLLMs~\cite{videomme, mvbench, tempcompass,hong2025worldsense}, a more in-depth and comprehensive evaluation has become more accessible. Compared to previous traditional benchmarks, these newly developed benchmarks are generally more challenging, often entailing composite skills and a finer-grained understanding of the video (\eg, the plot in the movie or causal relationships between events), and can be much longer in duration (\eg, up to 60 minutes in the Video-MME benchmark). 
In this work, our study adopts these newly developed video benchmarks.

\subsection{Textual Data for Video Understanding}
Since MLLMs are typically built upon LLMs and thus highly compatible with textual data, some works have explored utilizing pure text data to boost the performance of video understanding. 
Below, we outline the key ideas of these methods and their differences from our method.
\begin{itemize}
    \item \textbf{Textual data for context expanding}. Previous works have explored utilizing textual data to expand the context window of base LLMs. Specifically, in order to facilitate long video understanding, LLaMA-VID~\cite{llama-vid} and LongVA~\cite{longva} incorporate long text data in fine-tuning and continued pre-training stages, respectively, to expand the context window of LLM backbones. 
    
    \textbf{\underline{Differences}}: In this work, we (1) adopt textual data as a data augmentation method and (2) use them in the vision form to accommodate the training format.

    \item \textbf{Synthetic textual data for video understanding}. This line of work investigates synthesizing textual data that simulates video QA data, aiming to transfer temporal reasoning capabilities from textual training. 
    More specifically, TOPA~\cite{topa} extracts textual captions and object-level information from video frames, while T3~\cite{t3} gathers similar information from multiple different images.

    \textbf{\underline{Differences}}: These two works seek to boost video understanding with synthetic data, while ours aims to enrich the instruction diversity of the training corpus. Moreover, our method does not require calling advanced LLM APIs to build data; instead, our method utilizes existing datasets.
\end{itemize}

%% file: problem_formulation.tex
\section{Problem Formulation}
We focus on mainstream MLLMs architectures~\cite{yin2024survey}, which typically consist of a vision encoder, a projector, and an LLM backbone. Given a video $\mathbf{V}$ downsampled to $T$ frames as $\mathbf{F} = \{{f_i}\}_{i=1}^T$, frame-level features are extracted via $\mathbf{E} = \{\mathrm{E}_i\}_{i=1}^T = \texttt{ViT}(\mathbf{F})$, where \texttt{ViT} denotes the vision encoder (e.g., a pre-trained model such as CLIP~\cite{clip}). Each $\mathrm{E}_i \in \mathbb{R}^{(H \times W) \times C}$ represents the visual tokens of the $i$-th frame, with $H$, $W$ being the spatial dimensions and $C$ the feature dimension.
These visual features are then projected into the text embedding space via a projector module, typically an MLP, yielding $\mathbf{\hat{E}} = \texttt{Proj}(\mathbf{E})$. The transformed vision features are subsequently concatenated with the text embeddings of a user query $\mathbf{Q}$ to form a joint token sequence $[\textrm{w}_{V};\textrm{w}_{T}]$.
This combined sequence is fed into the LLM backbone, which generates a natural language response auto-regressively:
\begin{equation}
p(w_o|\textrm{w}_{V},\textrm{w}_{T}) \sim \prod_{t=1}^{L} P(w_t | w_{<t},\textrm{w}_{V},\textrm{w}_{T}),
\end{equation}
where $\textrm{w}_o=\{w_{o,t}\}_{t=1}^{L}$ denotes the output token sequence of length $L$. Here, $\textrm{w}_{V}$ corresponds to the vision tokens processed through the encoder and projector, while $\textrm{w}_{T}$ refers to the tokenized representation of the user query.

In terms of training data composition, each instance is structured as a triplet $(\mathbf{V}, \mathbf{Q}, \mathbf{A})$, where $\mathbf{Q}$ denotes a natural language instruction and $\mathbf{A}$ is the corresponding textual answer. Let $\mathcal{D}$ represent the data distribution over such triplets. The model is trained to generate the answer sequence $\mathbf{A}$ conditioned on both the video input $\mathbf{V}$ and the instruction $\mathbf{Q}$. Formally, we aim to minimize the expected negative log-likelihood loss over the model answer:
\begin{equation}
\mathcal{L}(\theta) = \mathbb{E}_{(\mathbf{V}, \mathbf{Q}, \mathbf{A}) \sim \mathcal{D}} \left[ -\log\, p_{\theta}(\mathbf{A} \mid \mathbf{V}, \mathbf{Q}) \right].
\end{equation}
where $\theta$ denotes the model parameters and $p_{\theta}(\cdot)$ is the conditional output distribution of the MLLM.

%% file: 03_motivation.tex
\section{A Probing Study of Data Scaling}
To understand the scaling characteristics of training data, our study starts with fine-tuning with different sample sizes and examines the relationship between training sample size and model performance.

In this section, we introduce the study's training and evaluation setup and then illustrate the empirical findings.

\subsection{Training Setup}

\subsubsection{Model Setup}
During our exploration, we mainly utilize two image-LLMs, including Mini-InternVL-Chat-4B-V1.5~\cite{internvl} (termed as InternVL hereafter), MiniCPM-Llama3-8B-V2.5~\cite{minicpm} (termed as MiniCPM-8B hereafter).
These instruction-tuned models are trained with massive image data and equipped with strong foundational capabilities. To support higher-resolution vision input, these models adopt the patchifying technique~\cite{ureader,monkey,sphinx} with a dynamic resolution scheme, where an image can be cropped into multiple sub-images according to different aspect ratios. Specifically, InternVL supports up to 13 sub-images, each of which is converted into 256 visual tokens; MiniCPM-8B slices images into a maximum of 10 patches, where each is represented by 96 visual tokens.
During training and evaluation, we switch off the patchifying option for higher efficiency.

\subsubsection{Training Configurations}
For fairness and ease of reproduction, we follow the official implementations. More specifically, we train the whole model end-to-end (except for InternVL-4B, where we freeze the vision encoder) with a learning rate of 5e-6.

\begin{figure}[!h]
\captionsetup[subfigure]{justification=centering}
	\centering
    \begin{subfigure}{0.5\columnwidth}
    \centering
        \includegraphics[width=\columnwidth]{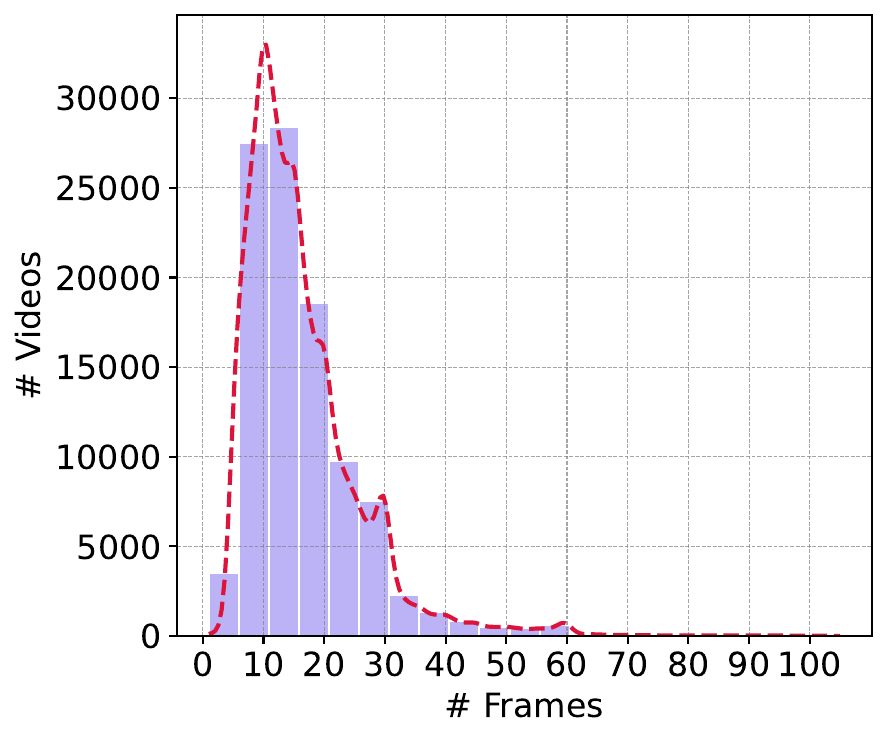}
        \caption{Share-Gemini}
    \end{subfigure}%
    \begin{subfigure}{0.5\columnwidth}
    \centering
        \includegraphics[width=\columnwidth]{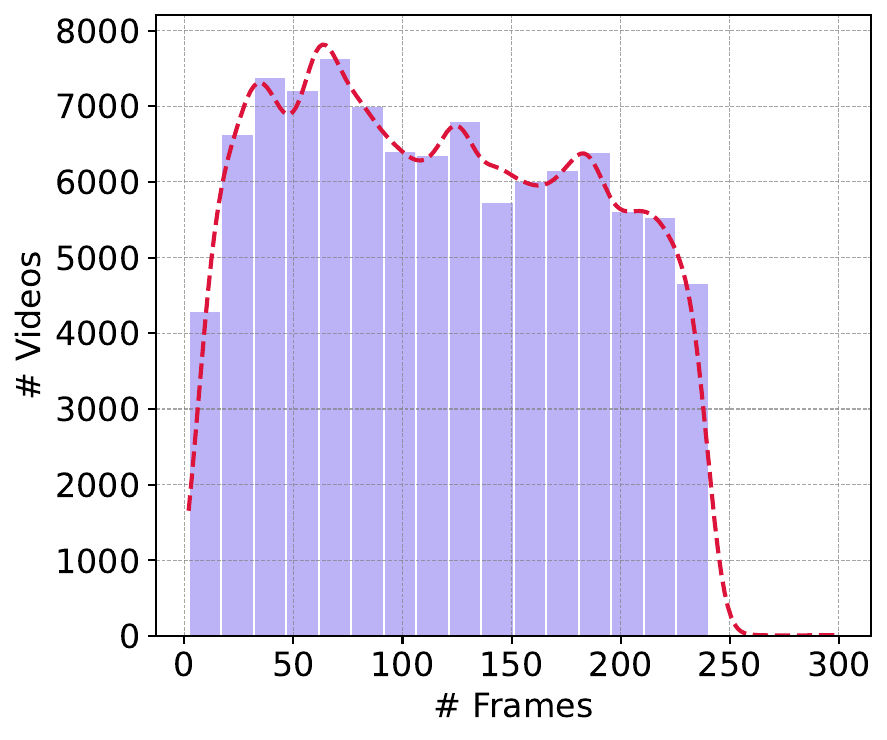}
        \caption{Video-ChatGPT}
    \end{subfigure}%
	\caption{\textbf{Video length statistics of ShareGemini and Video-ChatGPT datasets.} Both datasets mostly cover videos shorter than 3.5 minutes. 
    We extract video frames at an FPS of 1 for each video.. For better visibility, we pick samples with frame numbers lower than 99.9 percentile for visualization.}
	\label{fig_data_length}
\end{figure}

\subsubsection{Training Datasets}
During our investigation, we utilize two representative types of datasets, \ie, video-caption pairs and video instruction data. Specifically, we choose the ShareGemini~\cite{sharegemini} dataset and the Video-ChatGPT~\cite{videochatgpt} dataset as caption and instruction data, respectively. 
For each video, frames are extracted at an FPS of 1. In consideration of efficiency, we use up to 64 frames for InternVL-4B and 24 frames for MiniCPM-8B. 
When the total number of frames exceeds the threshold, we uniformly downsample the video frames. 
The statistics of video lengths are shown in~\Cref{fig_data_length}, and we provide more introduction to the two datasets below.

\noindent \textbf{ShareGemini-Webvid-core100k.}
This dataset comprises 100K video-caption pairs in total. The videos in the dataset are sourced from WebVid~\cite{webvid}, a web-scale video-caption dataset spanning diverse open-domain topics. In terms of temporal duration, the majority of videos are short-form, with lengths under 30 seconds.

The captions are annotated by calling the strong Gemini-1.5-Pro~\cite{gemini1dot5} API. To ensure the diversity of video content, an advanced clustering algorithm~\cite{tome} is used to filter out highly similar videos.
For simplicity, we refer to this dataset as ShareGemini in the following parts of the paper.

\noindent \textbf{Video-ChatGPT.}
The video instruction dataset contains 100K video-instruction pairs. The videos in this collection are derived from ActivityNet~\cite{caba2015activitynet}. The dataset's coverage of video duration is larger, yet the average video length is no more than 3.5 minutes. There are broadly three types of instructions: video summarization, questions about video content, and creative/generative tasks.

The dataset is annotated in a semi-automatic manner. A small portion of data samples are manually annotated by human annotators by refining and enriching the video captions. Other instruction data are generated by GPT-3.5 with the aid of off-the-shelf dense prediction and captioning models.

\subsection{Evaluation Setup}
To evaluate the model capabilities in an efficient and comprehensive way, we use Video-MME~\cite{videomme}, MVBench~\cite{mvbench}, and TempCompass~\cite{tempcompass} as our benchmarks. 
We do not use traditional video-QA benchmarks (\eg, MSVD-QA~\cite{msvd-qa}, TGIF-QA~\cite{tgif-qa}, ActivityNet-QA~\cite{activity-net-qa}) since these benchmarks are generally limited to a small coverage of domains, task types, and video lengths.
Moreover, the questions asked often involve shallow perception without deeper reasoning since early models generally lack reasoning capacity, whereas recent LLM-based models excel. 
We illustrate more about the benchmarks used as follows:

\noindent \textbf{Video-MME} is a comprehensive benchmark designed for the evaluation of video-LLMs. For temporal coverage, videos of short length (up to 2 minutes), medium length (4–15 minutes), and longer duration (30–60 minutes) are included.
The videos and annotations are manually collected and filtered.
We only use the raw frames without the subtitles to focus on the evaluation of video understanding capabilities.

\noindent \textbf{MVBench} designs a set of 20 video tasks that cover both perception and cognition, such as scene transition and episodic reasoning. Compared to Video-MME, the videos are sourced from existing benchmarks, and the QAs are automatically generated for the 20 pre-defined tasks.

\noindent \textbf{TempCompass} is designed to evaluate models' temporal understanding capabilities, encompassing temporal aspects such as action, speed, and attribute change. The videos and corresponding meta-information are manually collected, followed by LLM-generated annotations. We use the multiple-choice QA (MCQ) format to align with other benchmarks.

To ensure robust and efficient judging of model answers, we use a combination of exact matching and LLM matching for assessment. More details about the implementation of this evaluation scheme are available in Appendix A.

\subsection{Main Findings}

\subsubsection{Low Learning Efficiency Issue}
\begin{figure*}[!thb]\centering
	\includegraphics[width=\textwidth]{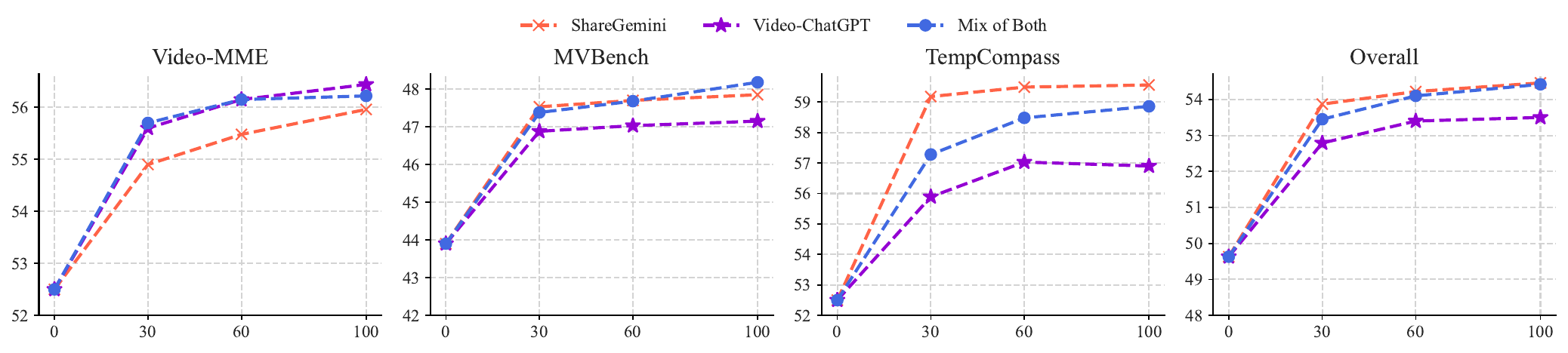}
	\caption{\textbf{Scaling performance of fine-tuning InternVL with different data volumes and types} on different benchmarks. We consider fine-tuning with ShareGemini, Video-ChatGPT, and a mix of both with a $1:1$ sampling ratio. Training samples are measured in K, where 0 sample indicates zero-shot inference.
    }
	\label{fig:scale_internvl}
\end{figure*}

\begin{figure}[!htbp]\centering
\setlength{\belowcaptionskip}{-4mm}
	\includegraphics[width=0.9\linewidth]{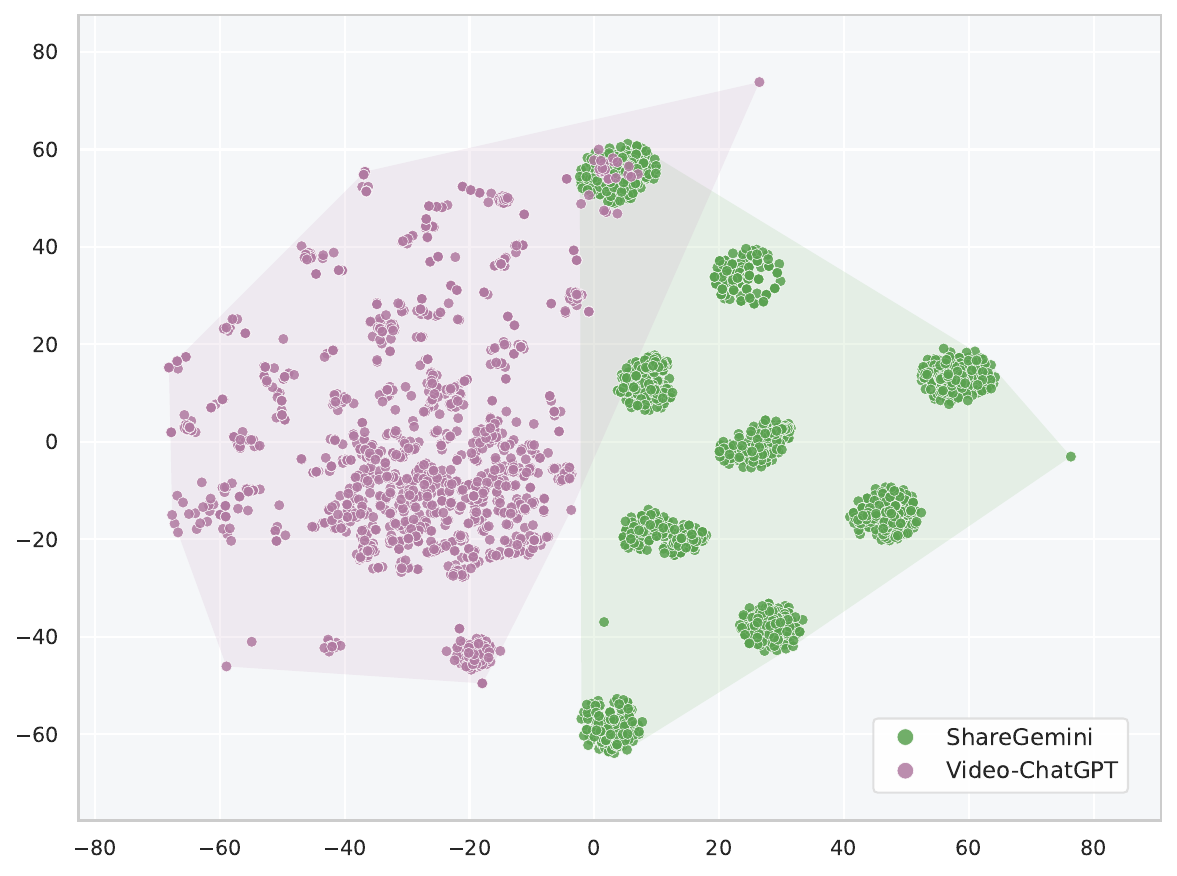}
	\caption{\textbf{t-SNE plot of instruction distributions for video datasets} – ShareGemini and Video-ChatGPT. We sample 5,000 instructions from each dataset for visualization. The relatively limited instruction diversity observed in both datasets may hinder learning efficiency during fine-tuning.}
	\label{fig:vis-instruction-distribution-video}
\end{figure}

Our experiments start with scaling up the training data volume and evaluating the video understanding performance on different general video understanding benchmarks. The results are shown in~\Cref{fig:scale_internvl}.
In general, training either with video caption data (ShareGemini), instruction data (Video-ChatGPT), or a mix of both can boost the image-LLM's video understanding performance. 
Meanwhile, increasing the training volume brings additional gains in accordance with the data scaling law.
However, the gains from scaling up quickly reach a plateau.
For instance, on the Video-MME benchmark, when training with mixed data, 30K samples improve overall accuracy by 3.1 points, while 100K samples only add another 0.5 points, resembling a logarithmic growth.  
In view of this quick and early saturation, the learning efficiency with these video datasets can be quite limited. 
The phenomenon also suggests that there could be high redundancy in the training corpus, and it is possible that we may use less data to achieve a performance comparable to or even better than training with more data samples.

\subsubsection{Probing of Instruction Diversity}

Previous results prompt us to explore the reason for such low learning efficiency.
Inspired by prior studies, which have underscored the importance of instruction diversity for fine-tuning LLMs~\cite{zhou2024lima} and image-LLMs~\cite{zeng2024matters}, we conduct an inspection of training data in this aspect.
Specifically, we follow previous approaches~\cite{magpie, zhao2024wildchat} to visualize the distribution of instructions in the training corpus. 5,000 instructions are sampled from ShareGemini and Video-ChatGPT, respectively. Then, the instructions are embedded and visualized using the t-SNE technique, as shown in~\Cref{fig:vis-instruction-distribution-video}.
\textbf{Overall, the instruction distribution of these two datasets is not diverse enough, which leads to a low data efficiency}: 
The distribution of ShareGemini exhibits 9 clear clusters in the figure, indicating very similar instructions. 
This is because this dataset samples from a fixed pool of 9 templates as instructions, each of which is a variant of ``Describe this video in detail''. 
On the other hand, the distribution of Video-ChatGPT seems relatively more diverse, as it includes specific questions related to video content and details besides video summarization. 
Nevertheless, the instruction diversity is still low due to the nature of self-instruction and a few fixed task-specific prompting templates for data curation.

%% file: 04_methods.tex
\section{Methods}

\subsection{Design Concept}
\begin{figure*}[!th]\centering
	\includegraphics[width=0.95\linewidth]{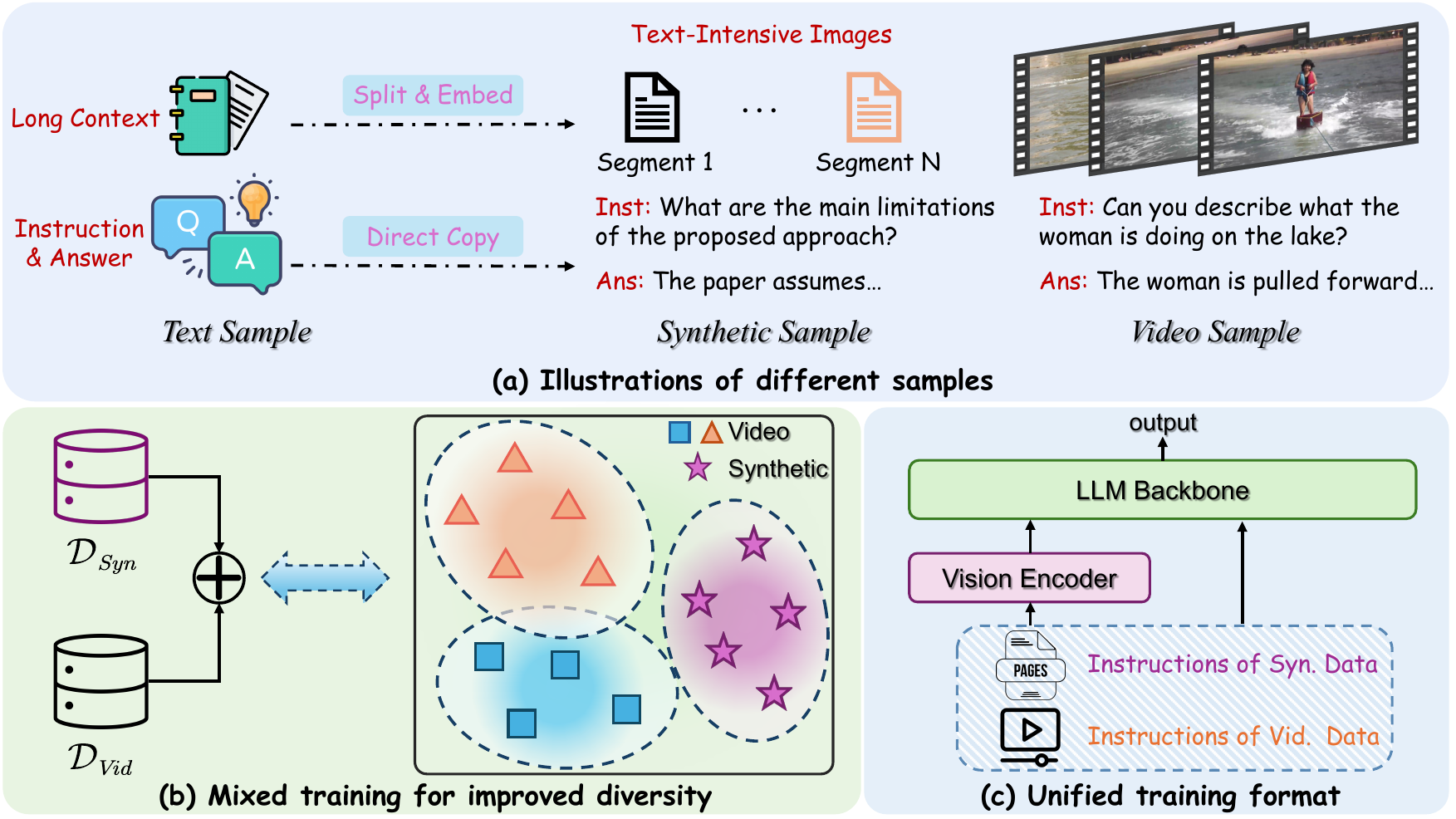}
    \caption{\textbf{Conceptual illustration of our proposed scheme.} 
(a) We illustrate the structures of text, synthetic, and video samples. Synthetic video-like samples are generated from textual data, with a structure mimicking real video samples. Specifically, for each (long context, instruction, answer) triplet, the long context part is split into segments, which are then transcribed into a sequence of text-rich images, simulating a video. 
(b) The domain differences between real and synthetic sources contribute to greater instruction diversity when used jointly for training. 
(c) The sequential nature of synthetic image sequences aligns well with the structure of real video data, thus enabling a unified training format.}
	\label{fig:idea}
\end{figure*}

\begin{figure}[!thbp]\centering
\setlength{\belowcaptionskip}{-3mm}
	\includegraphics[width=\columnwidth]{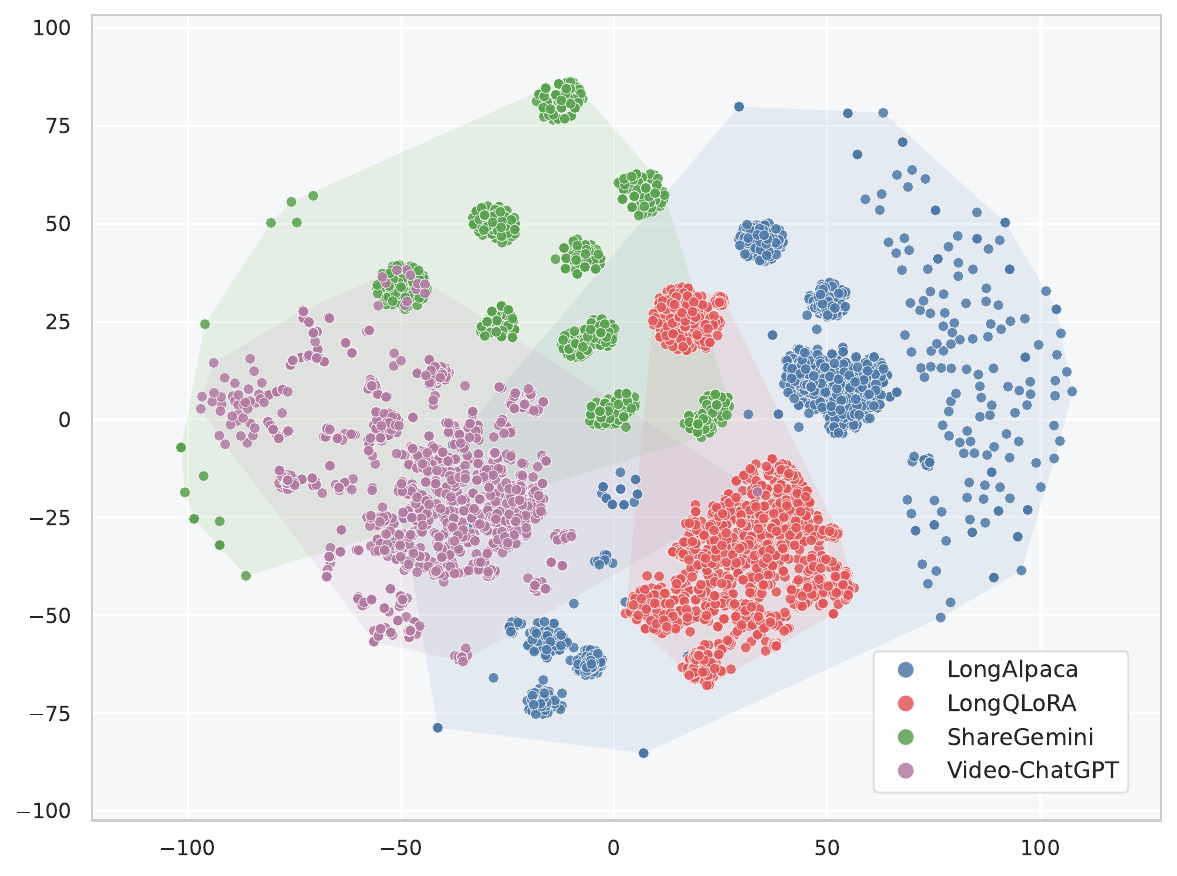}
	\caption{\textbf{t-SNE plot of instruction distribution after applying our proposed method.}}
	\label{fig:vis-instruction-distribution-mix}
\end{figure}

Since currently available video data can be limited in instruction diversity, and annotating high-quality video data is costly, we aim to expand the instruction diversity by incorporating new synthetic data. 
A rich source of instruction data lies in the text domain, and it can effectively complement the vision domain. Nevertheless, there is inherently a modality gap between the text and visual domains. To better utilize these data, we bridge the modality gap by synthesizing images with the text. \Cref{fig:idea} illustrates the overall data synthesis pipeline and characteristics of our scheme.
Our proposed scheme enjoys three benefits: 
(1) Mixing in text data can effectively enrich the instruction diversity (\Cref{fig:vis-instruction-distribution-mix}), thus improving the learning efficiency for video fine-tuning;
(2) Images synthesized from text can emulate the 1D temporal structure of video frames since text segments are generally correlated in the context, thus mitigating the gap between common video samples and synthetic ones;
(3) Text data is easier to collect than video samples. Thus, utilizing synthetic data can be economical.

\subsection{Implementation Details} 
Each text sample is a (long-context, question, answer) triplet. 
For example, the long context can be a section of a book or an academic paper, while the instruction and the answer are centered around the context, \eg, an inquiry to give a synopsis, or questions related to the paper.
After the data transformation process, each sample is a video-like (images, question, answer) triplet, where long-context information is transformed into a series of images, and the question and answer stay unchanged.

The key to the data synthesis procedure is synthesizing images with pure text. 
Specifically, for each (long-context, question, answer) triplet, we divide the context information into multiple segments according to word counts (set to 115 empirically) using an open-sourced NLP toolkit\footnote{\url{https://www.nltk.org/}}.
These text chunks are then transformed into a sequence of images separately. 
Specifically, each text chunk is embedded into a blank image with a white background.
This is achieved using a bitmap font with the ImageFont module of the Pillow library\footnote{\url{https://pillow.readthedocs.io/en/stable/}}. 
Each image is 448x448 pixels in size, and the font is 20 pt large, black color, \texttt{Arial Regular} type. 
We use a bounding box to control the layout, leaving a margin of 20 pixels on each side, so each line of the text has roughly the same width.
Following the transformation, the synthetic data aligns structurally with the video samples, enabling its direct and seamless incorporation into the video training sets.

%% file: 05_experiment.tex
\section{Evaluation on Proposed Methods}
This section includes experimental results and discussions of our proposed method, including (1) a comparison with mainstream methods, (2) an ablation study on data mixes, (3) an examination of key properties, including data scaling performance and gains in long video understanding, and (4) an in-depth analysis of model performance, including qualitative results as well as performance breakdown across various task types. 

\begin{table*}[!htbp]
\centering
\small
\renewcommand{\arraystretch}{1.2}
\setlength{\arrayrulewidth}{1.5\arrayrulewidth}
\setlength{\tabcolsep}{3mm}{
\begin{tabular}{lcccccc} \toprule
\textbf{Methods} & \textbf{Size} & \textbf{Frames} & \textbf{Short} & \textbf{Medium} & \textbf{Long} & \textbf{Overall} \\
\midrule
\rowcolor{gray!30}
\textsc{Proprietary Models} &&&&&&\\
GPT-4V~\cite{gpt-4v} & N/A & 10 & 70.5 & 55.8 & 53.5 & 59.9 \\
Claude 3.5 Sonnet~\cite{claude-3.5-sonnet} & N/A & 20 & 71.0 & 57.4 & 51.2 & 60.0 \\
GPT-4o~\cite{gpt-4o} & N/A & 384 & 80.0 & 70.3 & 65.3 & 71.9 \\
Gemini 1.5 Pro~\cite{gemini-1.5-pro} & N/A & 1fps & \textbf{81.7} & \textbf{74.3} & \textbf{67.4} & \textbf{75.0} \\ 
\midrule[0.05pt]
\midrule[0.05pt]

\rowcolor{gray!30}
\textsc{Open-Source Models} &&&&&&\\
VideoChat2~\cite{mvbench}         & 7B & 16 & 48.3 & 37.0 & 33.2 & 39.5 \\
Video-LLaVA~\cite{video-llava}        & 7B & 8 & 45.3 & 38.0 & 36.2 & 39.9 \\
Chat-UniVi-v1.5~\cite{chatuniv}    & 7B & 64 & 45.7 & 40.3 & 35.8 & 40.6 \\
VideoLLaMA 2~\cite{videollama2}       & 7B & 16 & 56.0 & 45.4 & 42.1 & 47.9 \\
VITA~\cite{vita}                       & 8x7B & 32 & 65.9 & 52.9 & 48.6 & 55.8 \\
Kangaroo~\cite{kangaroo}           & 8B  & 64 & 66.1 & \textbf{55.3} & 46.6 & 56.0 \\
VITA-1.5~\cite{fu2025vita} & 7B  & 16 & 67.0 & 54.2 & 47.1 & 56.1 \\

\midrule[0.01pt]

\rowcolor{gray!20}
\textsc{FT w/ InternVL}~\cite{internvl} &&&&&& \\
\rowcolor{gray!4}
\textcolor{gray}{Zero-shot}           & \textcolor{gray}{3.8B} & \textcolor{gray}{64} & \textcolor{gray}{61.3} & \textcolor{gray}{51.8} & \textcolor{gray}{44.3} & \textcolor{gray}{52.5} \\
\rowcolor{gray!8}
\textcolor{darkgray!85}{200K video data}           & \textcolor{darkgray!85}{3.8B} & \textcolor{darkgray!85}{64} & \textcolor{darkgray!85}{66.7} & \textcolor{darkgray!85}{54.2} & \textcolor{darkgray!85}{48.1} & \textcolor{darkgray!85}{56.3} \\
\rowcolor{LightBlue}
\textbf{Sparrow} (30K hybrid data)           & 3.8B & 64 & \textbf{67.0} & 53.7 & \textbf{49.3} & \textbf{56.7} \\

\bottomrule
\end{tabular}
}
\caption{\textbf{Accuracy comparisons of different methods on the Video-MME benchmark.} Performance is ranked in ascending order regarding overall performance (The models are grouped according to open-source or not). 
\colorbox{LightBlue}{Our method} uses only \textbf{15\%} of the total sample size compared to the full volume video datasets (200K) for fine-tuning and achieves comparable performance. 
30K hybrid data comprise 20K data sampled from video datasets and 10K synthesized from our method.
\textbf{Bold digits} indicate the best performance within each group.}
\label{tab:overall-compare}
\end{table*}

\subsection{Comparison with Mainstream Methods}

We compare the our method with some representative proprietary models, including GPT-4V~\cite{gpt-4v}, Claude 3.5 Sonnet~\cite{claude-3.5-sonnet}, GPT-4o~\cite{gpt-4o}, Gemini 1.5 Pro~\cite{gemini-1.5-pro}, and open-source video-LLMs of similar LLM parameter size, including Video-LLaVA~\cite{video-llava}, VideoChat2~\cite{mvbench}, Chat-UniVi-v1.5~\cite{chatuniv}, VideoLLaMA 2~\cite{videollama2}, VITA~\cite{vita}, VITA-1.5~\cite{fu2025vita}, and Kangaroo~\cite{kangaroo}. The results are summarized in~\Cref{tab:overall-compare}.

The table results show that, through zero-shot inference, the image-LLM Intern-VL already outperforms a variety of video-LLMs with larger LLM parameter sizes. 
This might be due to the rich pre-trained knowledge embedded in the model parameters since the image-LLM has been trained with large-scale and high-quality image-text data. 
The vision prior lays a strong foundation for further video fine-tuning, where models learn temporal and causal concepts from activities, events, \etc.
The model fine-tuned with full video datasets achieves an overall gain of 3.8 points on the image-LLM, closing the gap between open-source models and proprietary ones.

Notably, our methods use only \textbf{15\%} of the total sample size compared to the full volume (200K) for fine-tuning and achieve comparable performance. This result suggests the high data efficiency of our proposed scheme since mixing in synthetic samples mitigates the low instruction diversity issue illustrated in the earlier section.

\subsection{Ablation on Different Data Compositions}

\begin{table}[!htbp]
\centering
\small
\begin{tabular}{p{2.8cm}cccc}
\toprule[1pt]
\textbf{Data Mix} & \textbf{S} & \textbf{M} & \textbf{L} & \textbf{Overall} \\
\toprule[0.5pt]
30K Share-Gemini               & 65.7       & 52.8       & 46.1       & 54.9             \\
\toprule[0.2pt]
30K Video-ChatGPT               & 66.3       & 53.0       & 47.3       & 55.6             \\
\toprule[0.2pt]
\makecell[cl]{15K Share-Gemini\\15K Video-ChatGPT}              & 66.2       & 53.3       & 47.4       & 55.7             \\
\toprule[0.2pt]
\rowcolor{LightBlue}
\makecell[cl]{10K Share-Gemini\\10K Video-ChatGPT\\10K synthetic}               & 67.0       & 53.7       & 49.3       & 56.7             \\
\toprule[0.2pt]
\makecell[cl]{10K Share-Gemini\\10K Video-ChatGPT\\10K pure text}              & 67.3       & 52.4       & 47.7       & 55.8             \\
\toprule[0.5pt]
\toprule[0.5pt]
\rowcolor{gray!8}
\textcolor{darkgray!85}{Zero-shot}        & \textcolor{darkgray!85}{61.3}       & \textcolor{darkgray!85}{51.8}       & \textcolor{darkgray!85}{44.3}       & \textcolor{darkgray!85}{52.5}             \\
\toprule[0.2pt]
\rowcolor{gray!8}
\textcolor{darkgray!85}{200K full data}        & \textcolor{darkgray!85}{66.7}       & \textcolor{darkgray!85}{54.2}       & \textcolor{darkgray!85}{48.1}       & \textcolor{darkgray!85}{56.3}             \\
\bottomrule[1pt]
\end{tabular}
\caption{\textbf{Results of different data compositions on the Video-MME benchmark.} \colorbox{LightBlue}{Our proposed scheme} achieves an overall performance superior to other data mixes of the same amount (30K) and even more data (200K).}
\label{tab:ablation-data-ratio}
\end{table}

In order to examine the impact of different data compositions and validate the effectiveness of the proposed method, we conduct an ablation study and construct the following settings with the same amount of total data samples:
\begin{itemize}
    \item 30K video samples from ShareGemini.
    \item 30K video samples from Video-ChatGPT.
    \item 15K video samples from ShareGemini and 15K from Video-ChatGPT, respectively.
    \item \textbf{Our proposed scheme}: 10K samples each from ShareGemini and Video-ChatGPT, plus 10K samples synthesized from text data (5K from LongAlpaca and 5K from LongQLora, respectively).
    \item Same video samples as above (20K in total), plus 10K samples of corresponding pure text data.
\end{itemize}

\noindent \textbf{Examination of our design choices.}
As shown in~\Cref{tab:ablation-data-ratio}, comparing the first three rows, we can find that when using the same amount of video samples, training only with ShareGemini is not as effective as using more diverse data compositions. 
Notably, under the same data budget, our proposed scheme (Row 4) achieves the best performance.
Furthermore, compared to the full fine-tuning setting with 200K samples, our method achieves comparable performance using only \textbf{15\%} of the number of training samples. The training cost is also significantly reduced from 276.8 GPU hours to 33.6 GPU hours, yielding an 8.2$\times$ improvement in efficiency. These results underscore the importance of instruction diversity and validate the effectiveness of our proposed approach.

Notably, replacing the synthetic data with the original pure text counterpart achieves an overall inferior performance. We hypothesize that this is due to the inherent domain gap between vision and text. Thus, to simulate the structure of video frame sequences, transcribing long text into images is necessary.

\begin{figure*}[!thbp]\centering
	\includegraphics[width=\linewidth]{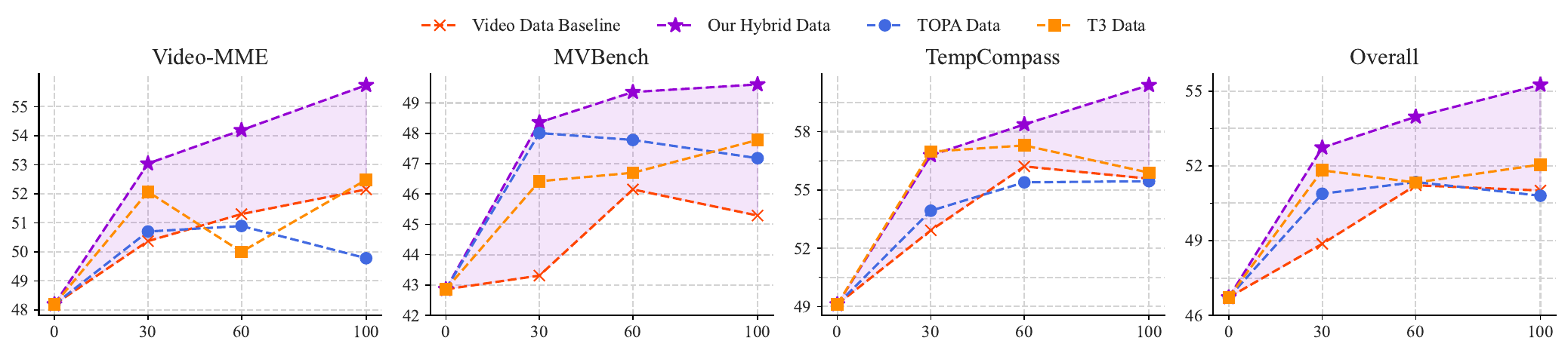}
	\caption{\textbf{Scaling performance with different data volumes and types on general video benchmarks}. Specifically, we scale up the volume of training data until the overall performance gain saturates (empirically set as 1 point). Four settings are considered, \ie, fine-tuning with video data only, our hybrid data (video data mixed with our synthetic samples with a ratio of $2:1$), TOPA data, and T3 data. By default, video data (the caption dataset and the instruction dataset) are mixed with a $1:1$ sampling ratio. Training samples are measured in K, where 0 sample indicates zero-shot inference. \textcolor{violet!60}{Purple-shaded} area indicates gains of our methods over the \textcolor{red!70}{video data baseline}.
    }
	\label{fig:scale_minicpm}
\end{figure*}

\noindent \textbf{Can synthetic data help models understand longer videos?}
Interestingly, in the training stage, we only utilize synthetic samples of long multimodal context instead of authentic long video samples. However, on the long video benchmark set, our proposed method still achieves a score that is 1.2 points higher than the full data training (as shown in the 4th row compared to 200K full data in \Cref{tab:ablation-data-ratio}). This result suggests that fine-tuning with a \textit{long multimodal context} can enhance the comprehension of longer videos. In the following section, we will present additional results and discussions to illustrate this point further.

\begin{table*}[!thbp]
\centering
\small
\setlength{\arrayrulewidth}{1.2\arrayrulewidth}
\setlength{\tabcolsep}{3.5mm}{
\begin{tabular}{lcccccc} \toprule
\textbf{Methods} & \textbf{Samples (K)} & \textbf{Frames} & \textbf{Video-MME$_\texttt{L}$} & \textbf{LongVideoBench} & \textbf{MLVU} & \textbf{Overall} \\
\midrule

\rowcolor{gray!4}
\multirow{7}{*}{\textsc{Baseline}} & \textcolor{gray}{0} & \textcolor{gray}{24} & \textcolor{gray}{40.1} & \textcolor{gray}{40.0} & \textcolor{gray}{44.5} & \textcolor{gray}{41.6} \\ 
 & 30 & 24 & 44.7 & 39.7 & 45.4 & 43.3 \\
 & 30 & 48 & 45.3 & 39.6 & 45.3 & 43.4 \\ \cmidrule[0.01pt]{2-7}
 & 60 & 24 & 46.2 & 42.7 & 46.2 & 45.1 \\
 & 60 & 48 & 46.2 & 42.1 & 45.0 & 44.5 \\ \cmidrule[0.01pt]{2-7}
& 100 & 24 & 46.7 & 44.1 & 45.3 & 45.3 \\
& 100 & 48 & 44.9 & 40.7 & 45.0 & 43.5 \\

\midrule[0.05pt]
\midrule[0.05pt]

\multirow{3}{*}{\textsc{Our Method}} & 30 & 24 & 45.6\plusvalue{0.9} & 48.7\plusvalue{9.0} & 51.4\plusvalue{6.0} & 48.5\plusvalue{5.2} \\
 & 60 & 24 & 46.2 & 51.2\plusvalue{8.5} & 53.2\plusvalue{7.0} & 50.2\plusvalue{5.1} \\
 & 100 & 24 & 48.7\plusvalue{2.0} & 50.1\plusvalue{6.0} & 57.0\plusvalue{11.7} & 51.9\plusvalue{6.6} \\

\bottomrule
\end{tabular}
}
\caption{\textbf{Long video understanding performance with different data volumes and types.} The \textsc{Baseline} method adopts only video data, while \textsc{Our Method} utilizes a mix of video data and synthetic samples attained from our method. 
In the \textsc{Baseline} group, 0 sample indicates zero-shot inference.
The \textbf{\textcolor{DarkGreen}{performance gains}} are calculated relative to the video data fine-tuning baseline trained with the same number of samples. 
}
\label{tab:long-evaluation}
\end{table*}

\subsection{Examination of Key Properties}
In this section, we further examine the key properties of our proposed method, including its general effectiveness, scaling performance, and effectiveness when applied to long video understanding scenarios.

\subsubsection{General Effectiveness and Scaling Performance}

We further verify the proposed scheme's effectiveness by evaluating our methods on another image-LLM of larger parameter size, \ie, MiniCPM-8B, across different benchmarks. 
Through scaling up with different volumes and types of data, we compare our methods against a pure video data baseline, as well as other relevant methods, including TOPA and T3. 
Both TOPA and T3 first translate vision information into text, such as captions and relations between objects.
Then, synthetic text QAs are constructed to simulate video reasoning samples, aiming to transfer temporal reasoning capabilities from text to video. 
Note that since the original data format of TOPA is largely different from the current paradigm, we design a template to adapt the samples to the instruction data format (More details are available in Appendix B). 
The results are summarized in~\Cref{fig:scale_minicpm}.

\noindent \textbf{General effectiveness.}
When using the same amount of training samples, our methods almost always outperform other methods by a clear margin in all the evaluated benchmarks. 
Specifically, when using 30K samples, our method achieves an overall accuracy of 52.7, surpassing the baseline by 3.9 points. Notably, it even outperforms the baseline trained with 100K samples by 1.7 points.
Similarly, on the MVBench benchmark, with 100K samples, our method attains a 4.3-point absolute gain over the baseline.
Overall, the superior performance on different benchmarks showcases the general effectiveness of our proposed method.

\noindent \textbf{Scaling performance.}
A notable limitation of other methods is that they are more prone to performance saturation when scaling up the data budget.
For instance, the baseline method uses 60K samples to improve the Video-MME benchmark by 3.1 points, while using 100K samples only achieves another 0.9 points of absolute gains.  
In contrast, our proposed scheme shows more stable and consistent improvements when scaling up the data volumes compared with other methods. 
This underscores the importance of maintaining instruction diversity in video-language training, as insufficient diversity can lead to reduced learning efficiency.
In such cases, our data augmentation strategy facilitates a more diverse instructional distribution.

\noindent \textbf{Discussion: Can we scale up only with synthetic textual samples?}
An intriguing and highly relevant question is whether we can scale up the synthetic samples without using any real video samples. Since the text is more compact and less redundant than a whole video, training in this way is more economical.
Unfortunately, the empirical results show that this is probably unfeasible.
As shown in~\Cref{fig:scale_minicpm}, scaling with synthetic text data (TOPA and T3) shows undesirable characteristics, \ie, this approach can easily reach the saturation point or even slightly downgrade.
Other critical issues include the modality gap and special processing of videos in various domains (such as egocentric videos and movies). 
Besides, since text suffers inevitably from information loss when translated from videos, text data might be better used as a supplement to videos, which helps inject temporal reasoning language prior into the LLM backbones (similar to TOPA or T3) or mixing with video data as a regularization method (as our method does).

\begin{figure}[!htbp]\centering
\setlength{\belowcaptionskip}{-4mm}
	\includegraphics[width=0.95\columnwidth]{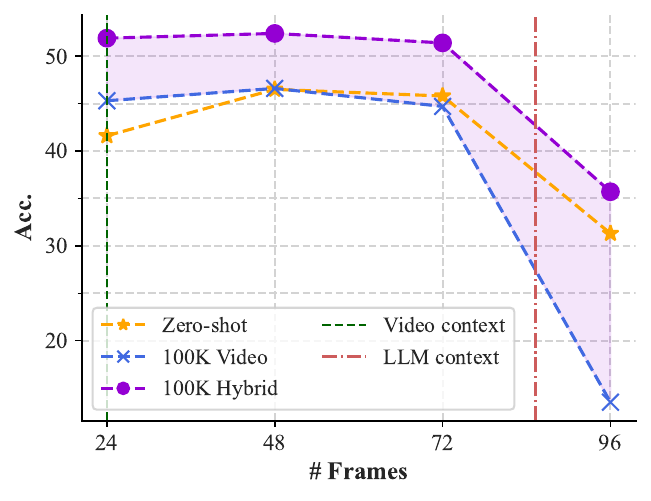}
	\caption{\textbf{Long video understanding performance evaluated with a different number of frames.}
    ``\textcolor{DarkGreen}{Video context}'' denotes the max number of video frames used in training, \ie, 24 frames; ``\textcolor{IndianRed}{LLM context}'' denotes the context window of the LLM backbone.
    \textcolor{violet!60}{Purple-shaded} area indicates performance gains of our proposed method over the \textcolor{RoyalBlue}{video data baseline}.}
	\label{fig:perf-frame-change}
\end{figure}

\subsubsection{Long Video Understanding Performance}

We adopt tailored benchmarks to evaluate long video understanding capabilities, including LongVideoBench~\cite{longvideobench}, MLVU-M~\cite{mlvu}, and the long video set of Video-MME, and report the performance on evaluation sets for the former two benchmarks.
Our study focuses on two aspects: (1) performance improvement in terms of long video understanding compared with the video fine-tuning baseline, and (2) frame number (multimodal context) generalization ability in the inference stage.

\noindent \textbf{Performance change \wrt training configurations.}
As presented in~\Cref{tab:long-evaluation}, our method consistently outperforms the video-only fine-tuning baseline in long video understanding across varying sample sizes, despite the absence of any long video training data.
Remarkably, with the same number of 100K training samples, our hybrid data strategy achieves a performance gain of 6.6 points over the baseline.
We attribute this improvement to the transferable reasoning patterns embedded in long-form textual data, which may enhance the model's temporal comprehension capabilities without requiring vision-derived textual supervision~\cite{topa,t3}.

We also examine whether adopting a denser sampling scheme can improve long video understanding performance, as this approach can also expand the context window. Specifically, we double the sampling frame limit (\ie, from 24 to 48) for authentic videos and find no improvements in long video understanding performance. This is because the training dataset predominantly contains short videos, and the dynamics within each video are relatively small. In this case, sampling more frames does not introduce more information but instead can bring redundancy, and thus, achieve no improvements in long video understanding.

\noindent \textbf{Performance change \wrt frame number.}
We also investigate whether our approach expands the context window.
Larger context windows are usually beneficial since video-LLMs derived from large-context LLM backbones often accompany frame number generalization and benefit from inputting more video frames~\cite{longvideobench, longva}.
However, we do not observe this trend.
As shown in~\Cref{fig:perf-frame-change}, the model fine-tuned with long synthetic samples still follows a similar pattern, that is, when performing inference beyond the video context of the training stage, the performance stays relatively stable and does not benefit from more frame input. And when the frame number exceeds the LLM context, the performance plunges to a low level. 
Thus, promising directions for further improvements may include continued pre-training to expand the LLM context window.

\begin{figure*}[!thbp]
\centering
\setlength{\belowcaptionskip}{-1mm}
	\includegraphics[width=0.95\linewidth]{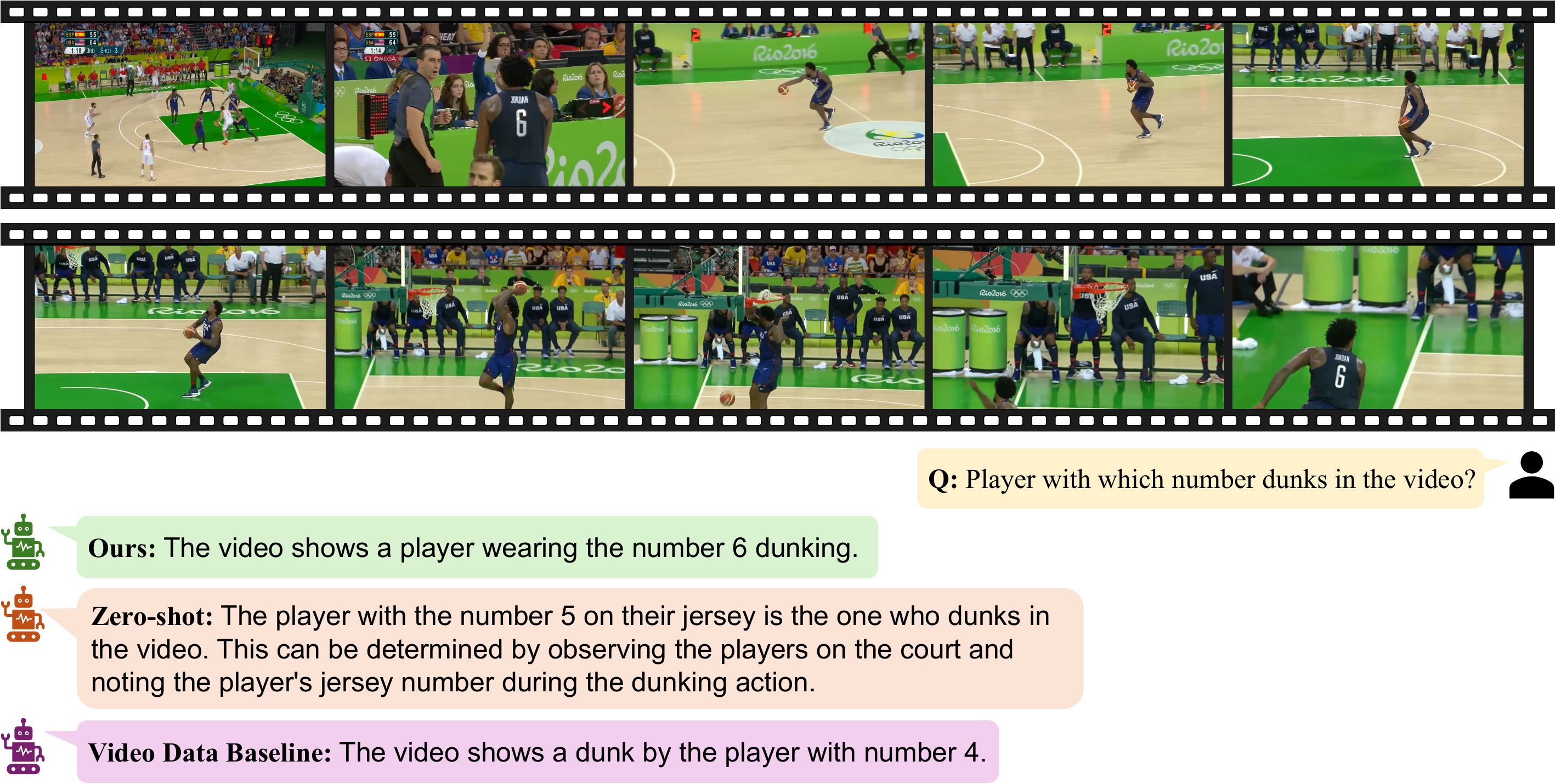}
     \caption{\textbf{Qualitative results of OCR-related capabilities.}
     The video features the process of dunking by a basketball player.
     Three settings are compared, \ie, zero-shot (direct inference with image-LLM), video data baseline (trained with 100K video samples), and our Sparrow scheme (trained with 66K video samples and 34K synthetic samples).
     }
	\label{fig:case-1}
\end{figure*}

\begin{figure}[!htbp]\centering
\setlength{\belowcaptionskip}{-3mm}
	\includegraphics[width=0.95\columnwidth]{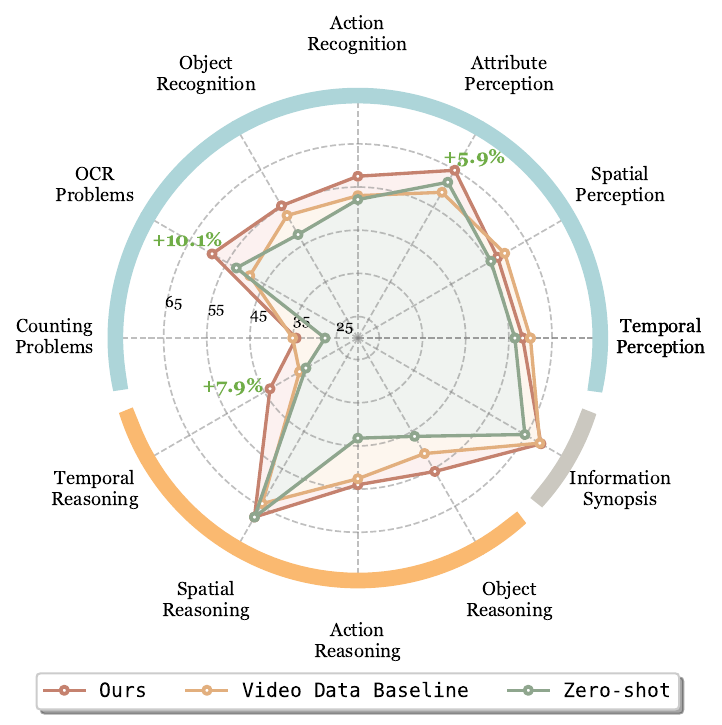}
	\caption{\textbf{Model performance across various task categories on Video-MME.}
The \textcolor{perception_color}{Perception} and \textcolor{cognition_color}{Cognition} categories are highlighted in the outer circle. Performance gains exceeding 5 points over the video data baseline are marked in \textcolor{mark_green}{green}.}
	\label{fig:radar-perf}
\end{figure}

\subsection{Fine-grained Performance Analysis}

In this section, we present a more in-depth analysis of model performance. By examining qualitative results and task-wise performance breakdowns, we provide intuitive insights into how our method influences model performance.

\subsubsection{Qualitative Results}

We present qualitative results on Video-MME in~\Cref{fig:case-1}, with additional examples provided in Appendix C. As shown in the figure, among the three models compared, only our method successfully identifies the number on the back of the basketball player, demonstrating superior OCR capabilities in video-based understanding. Further analyses on this aspect are discussed in the following section.

\subsubsection{Performance Breakdown}
To understand how our method influences model capabilities, we break down the performance across different tasks in Video-MME~\cite{videomme}.
Three broad categories of specific capabilities involved in video understanding are considered:

\noindent \textbf{Perception} includes Counting Problems, OCR Problems, Object Recognition, Action Recognition, Attribute Perception, Spatial Perception, and Temporal Perception.

\noindent \textbf{Cognition} includes Temporal Reasoning, Spatial Reasoning, Action Reasoning, and Object Reasoning.

\noindent \textbf{Information Synopsis.}
This category of questions emphasizes a model’s ability to comprehend the overall narrative or theme of a video rather than fine-grained details.

As shown in~\Cref{fig:radar-perf}, beyond overall gains, the incorporation of synthetic data yields notable improvements in OCR-related tasks. In particular, the model trained with augmented synthetic samples outperforms the one trained on 100K real video samples by 10.4\% in OCR problems.
This notable gain aligns with expectations, as synthetic data enables the model to more effectively recognize and interpret textual elements embedded in text-intensive frames, thereby enhancing its ability to extract relevant information.
Interestingly, a significant improvement is also observed in temporal reasoning, with a 7.9\% performance gain. This observation suggests that the ability to comprehend textual context may partially transfer to the modeling of temporal relationships across multiple video frames.


%% file: 10_conclusion.tex
\section{Conclusion}
\label{sec:conclusion}
In this paper, we propose \textbf{Sparrow}, a data-efficient training scheme for video-LLMs, which enables training with fewer samples and achieving better video understanding performance.
This method derives from our empirical findings that the low learning efficiency in data scaling may be ascribed to a limited instruction diversity in the training corpus.
Thus, we design an economical data augmentation method that synthesizes video-like samples rich in instruction diversity.
Comprehensive experiments demonstrate the general effectiveness and key properties of our proposed method. 
We hope this paper's findings can spark more explorations of efficient training and high-quality video training corpora.

%% file: 12_appendix.tex
\section{Answer Judging}
We notice that MiniCPM-8B~\cite{minicpm} often fails to follow instructions properly when we explicitly ask the model to ``\texttt{Answer with the option's letter from the given choices directly}'', making simple exact matching inaccurate. Specifically, the model often prepends or appends additional text other than the option letters, \eg ``\texttt{Answer: B. Pink.}'', or gives additional explanations apart from the answer.

To cope with these issues, we adopt a combination of exact matching and LLM matching for assessment. Specifically, we strip the prefixes such as ``\texttt{Answer:}'' from the prediction and try to use regular expression matching to find the option letter. When the exact matching scheme fails, we use an LLM (Llama-3.1-8B-Instruct~\cite{llama}) to find an option closest to the model prediction. When the LLM matching fails, a placeholder outside of the available options (such as ``Z'') is returned to denote a wrong answer. Our judging prompt for the LLM is modified from VLMEvalKit~\cite{vlmevalkit}, as shown in~\Cref{tab:eval_prompt}.

\section{Reproduction Details of Baseline Methods}
Due to an inconsistent formulation between the method TOPA~\cite{topa} and our proposed method, we adapt the implementation for a fair comparison.
The original sample comprises a global caption for the whole video and frame-specific information. The frame-related information contains a frame-level caption and some descriptions of key objects in the frame.
Thus, we design a prompt template to fit the original textual samples into the unified training format. 

A real case of formatting the sample with the devised template is shown in~\Cref{tab:train_prompt}. 

\begin{table*}[thbp!]\centering
\begin{minipage}{0.95\textwidth}    
\centering
\begin{tcolorbox} 
    \centering
      \small
    \begin{tabular}{p{0.95\textwidth}}
   \VarSty{ {\bf System message} } \\
You are an AI assistant who will help me match an answer with several options of a single-choice question. \\
    \midrule
   \VarSty{ {\bf Prompt} } \\
You are provided with a question, several options, and an answer, and you need to find which option is most similar to the answer. \\
If the meaning of all options is significantly different from the answer, output Z. You should directly output a single uppercase character, such as A, B, C, D (if they are valid options), and Z, and nothing else. Here are two examples.\\\\
Example 1: \\
Question: What is the main object in the image?\\\\Options: A. teddy bear.\\B. rabbit.\\C. cat.\\D. dog.\\
Answer: a cute teddy bear\\Output: A\\\\
Example 2: \\
Question: What is the main object in the image?\\Options: A. teddy bear.\\B. rabbit.\\C. cat.\\D. dog.\\
Answer: Spider\\Output: Z\\\\
Now here are the question, options, and the answer, you should match and give me the option letter: \\
Question: \textcolor[rgb]{0,0.7,0}{ \{Question\} }\\Options: \textcolor[rgb]{0.8,0,0}{\{Options\}}\\Answer: \textcolor[rgb]{0,0,0.8}{\{Model Answer\}}\\Output: 
    \end{tabular}
\end{tcolorbox}
\caption{Template for prompting LLM to perform option matching. \textcolor[rgb]{0,0.7,0}{ \{Question\} } is the specific question of a benchmark sample, and \textcolor[rgb]{0.8,0,0}{\{Options\}} are corresponding choices of the question. \textcolor[rgb]{0,0,0.8}{\{Model Answer\}} is the raw prediction of MLLMs.}
    \label{tab:eval_prompt}
\end{minipage}
\end{table*}

\begin{table*}[thbp!]\centering
\begin{minipage}{0.95\textwidth}    
\centering
\begin{tcolorbox} 
    \centering
      \small
    \begin{tabular}{p{0.95\textwidth}}
   \VarSty{ {\bf User:} } \\
You will be provided with some information about a video, including a global caption for the whole video, global captions for each video frame, and descriptions of key objects in each frame.
Answer the questions using the information below. \\\\

Video information:\\
Caption: This video shows the carpentry process. At first, the person sits on a workbench and measures the length of the plank. Then, he uses a saw to cut the wooden plank into multiple pieces. Then, he uses a hammer to nail two pieces of wood together. Finally, he takes a break and drinks water, and leaves the workshop.\\\\

Frame information:\\
Frame 1:\\
Caption: the person sits on a workbench and holds a hammer in his right hand.\\
the person: the person is sitting on a workbench in the center of the frame. He is looking at a wooden plank in front of him.\\
hammer: He has a hammer in his right hand.\\
workbench: He is sitting on a workbench in the center of the frame.\\\\

Frame 2:\\
Caption: the person takes a chisel from a red toolbox and starts chiseling the plank.\\
chisel: There are a chisel and a hammer on the workbench in the center of the frame.\\
the person: the person is holding a chisel in the center of the frame.\\
red toolbox: There is a red toolbox in the right corner of the frame.\\
workbench: He is sitting on a workbench in the center of the frame.\\\\

Frame 3:\\
Caption: the person measures the length of the plank with a measuring tape.\\
the person: the person carries a pencil and a ruler in his right hand.\\
pencil: There is a pencil in the right hand of the person.\\
measuring tape: the person is holding a measuring tape in his right hand.\\
wooden plank: The plank is above the workbench.\\\\

...\\\\

Frame 8:\\
Caption: the person puts his hammer and chisel into the toolbox and leaves the workshop.\\
toolbox: the person is putting a hammer and a chisel into a toolbox in the center of the frame.\\
the person: the person is walking toward the door in the center of the frame.\\
hammer: There are a hammer and a chisel in the right hand of the person.\\
workshop: the person is leaving the workshop.\\\\

Question: Why does the person use a measuring tape in the third frame?\\
A. To hammer the plank\\
B. To cut the plank\\
C. To measure the length of the plank\\
D. To handle the saw\\
E. To fix the length of the wooden plank\\
Answer with the option's letter from the given choices directly.\\

\midrule

\VarSty{ {\bf Assistant:} } \\
C
    \end{tabular}
\end{tcolorbox}
\caption{An example of using the template for structuring textual samples into the training format. For simplicity, we only show part of the frame-level information.}
    \label{tab:train_prompt}
\end{minipage}
\end{table*}

\section{More Qualitative Results}

\begin{figure*}[!th]
\centering
\includegraphics[width=0.95\linewidth]{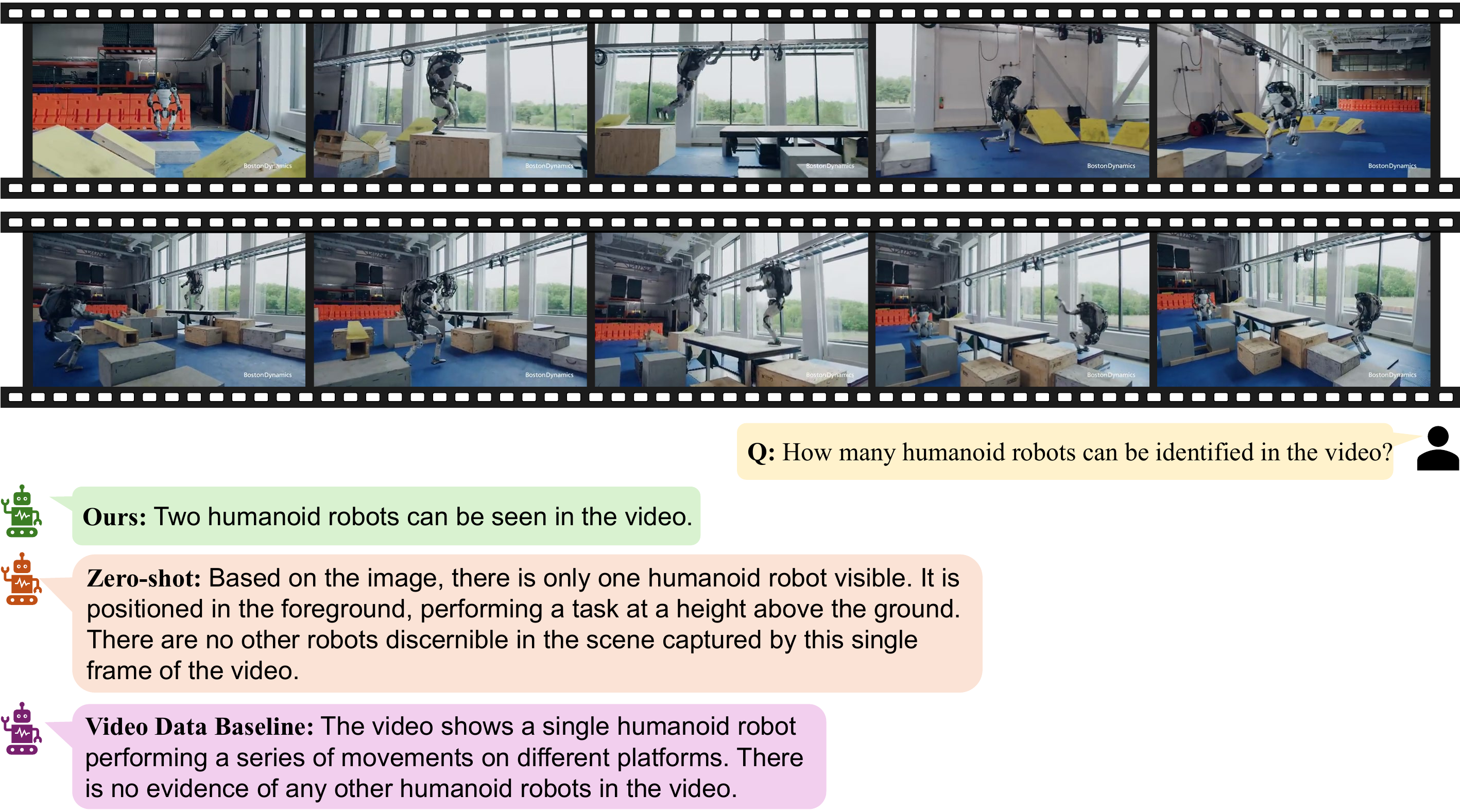}
     \caption{\textbf{A qualitative example from Video-MME.}
     The video depicts humanoid robots performing a sequence of actions.
     Our model demonstrates a clear understanding of the video content by accurately detecting both motion patterns and the presence of two distinct robots.}
	\label{fig:case-2}
\end{figure*}

\begin{figure*}[!thbp]
\centering
\includegraphics[width=0.95\linewidth]{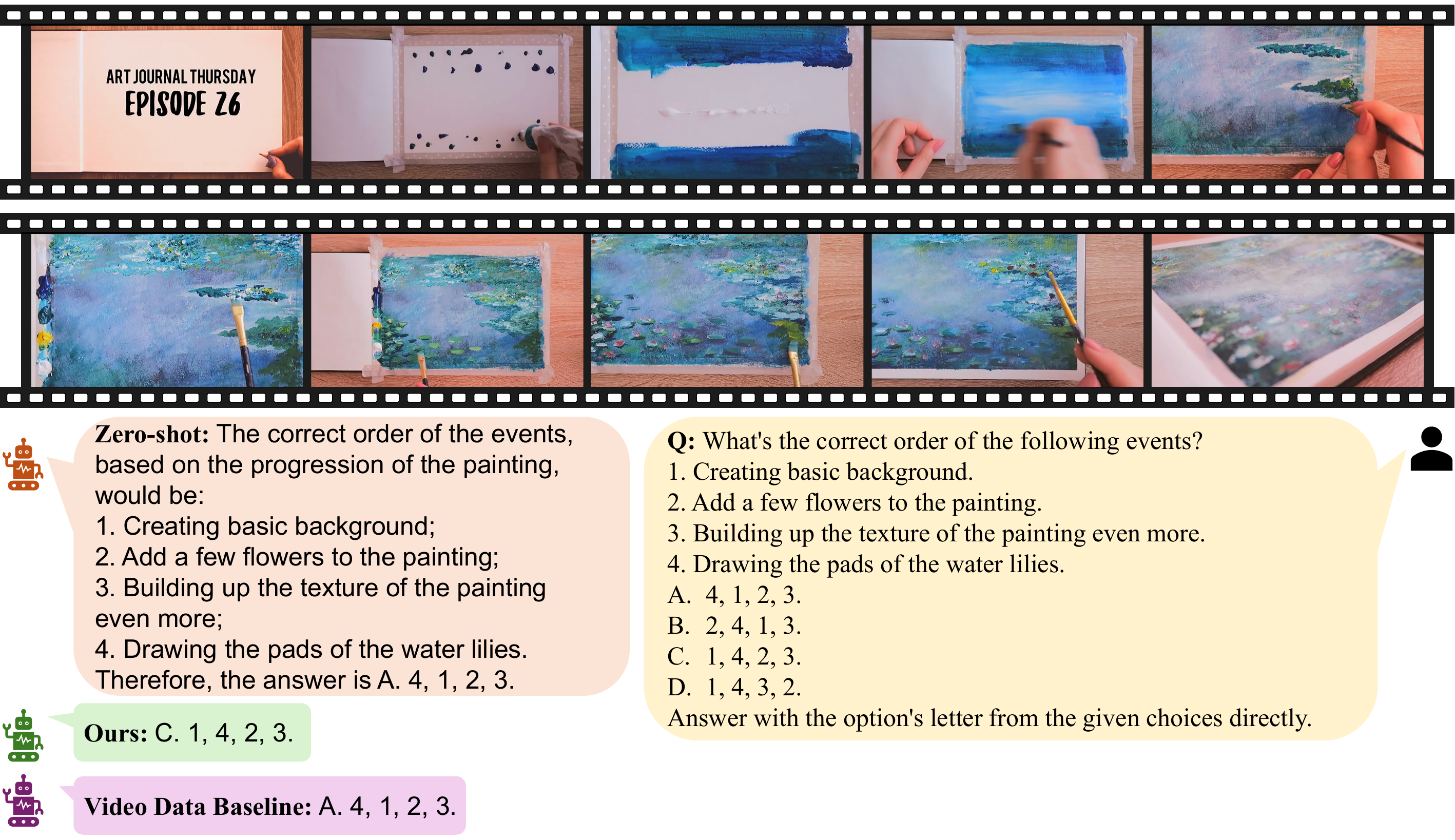}
     \caption{\textbf{A qualitative example from Video-MME.}
The video presents a tutorial on watercolor painting. In the zero-shot setting, the model exhibits an incorrect reasoning process, ultimately producing a conclusion that contradicts its own reasoning chain. In contrast, only our model demonstrates strong temporal reasoning abilities, successfully arriving at the correct answer.}
	\label{fig:case-3}
\end{figure*}

In this section, we present more qualitative results, as illustrated in~\Cref{fig:case-2} and~\Cref{fig:case-3}.

%% file: _main.bbl
\begin{thebibliography}{65}
\providecommand{\natexlab}[1]{#1}
\providecommand{\url}[1]{\texttt{#1}}
\expandafter\ifx\csname urlstyle\endcsname\relax
  \providecommand{\doi}[1]{doi: #1}\else
  \providecommand{\doi}{doi: \begingroup \urlstyle{rm}\Url}\fi

\bibitem[Yao et~al.(2024)Yao, Yu, Zhang, Wang, Cui, Zhu, Cai, Li, Zhao, He, et~al.]{minicpm}
Yuan Yao, Tianyu Yu, Ao~Zhang, Chongyi Wang, Junbo Cui, Hongji Zhu, Tianchi Cai, Haoyu Li, Weilin Zhao, Zhihui He, et~al.
\newblock Minicpm-v: A gpt-4v level mllm on your phone.
\newblock \emph{arXiv:2408.01800}, 2024.

\bibitem[Yin et~al.(2024)Yin, Fu, Zhao, Li, Sun, Xu, and Chen]{yin2024survey}
Shukang Yin, Chaoyou Fu, Sirui Zhao, Ke~Li, Xing Sun, Tong Xu, and Enhong Chen.
\newblock A survey on multimodal large language models.
\newblock \emph{National Science Review}, 2024.

\bibitem[Xiong et~al.(2025)Xiong, Zhuge, Zhu, Zhang, and Lu]{xiong20253ur}
Haomiao Xiong, Yunzhi Zhuge, Jiawen Zhu, Lu~Zhang, and Huchuan Lu.
\newblock 3ur-llm: An end-to-end multimodal large language model for 3d scene understanding.
\newblock \emph{IEEE Trans. Multimedia}, 27:\penalty0 2899--2911, 2025.

\bibitem[Wang et~al.(2025)Wang, Tian, Wang, and Yang]{wang2025multimodal}
Binglu Wang, Yao Tian, Shunzhou Wang, and Le~Yang.
\newblock Multimodal large models are effective action anticipators.
\newblock \emph{IEEE Trans. Multimedia}, 27:\penalty0 2949--2960, 2025.

\bibitem[Li et~al.(2024{\natexlab{a}})Li, Hu, Chen, Ma, Xu, and Zhang]{li2024lmeye}
Yunxin Li, Baotian Hu, Xinyu Chen, Lin Ma, Yong Xu, and Min Zhang.
\newblock Lmeye: An interactive perception network for large language models.
\newblock \emph{IEEE Trans. Multimedia}, 26:\penalty0 10952--10964, 2024{\natexlab{a}}.

\bibitem[Cao et~al.(2025)Cao, Zhao, Hao, Chai, Hwang, Wang, and Wang]{cao2025efficient}
Shidong Cao, Zhonghan Zhao, Shengyu Hao, Wenhao Chai, Jenq-Neng Hwang, Hongwei Wang, and Gaoang Wang.
\newblock Efficient transfer from image-based large multimodal models to video tasks.
\newblock \emph{IEEE Trans. Multimedia}, 27:\penalty0 3045--3056, 2025.

\bibitem[Yin et~al.(2025)Yin, Chen, Zhang, Jiang, Zhao, Liu, Yu, and Chen]{yin2025shapegpt}
Fukun Yin, Xin Chen, Chi Zhang, Biao Jiang, Zibo Zhao, Wen Liu, Gang Yu, and Tao Chen.
\newblock Shapegpt: 3d shape generation with a unified multi-modal language model.
\newblock \emph{IEEE Trans. Multimedia}, 27:\penalty0 4107--4120, 2025.

\bibitem[Liu et~al.(2025)Liu, Miao, Cao, Zhu, Ge, Liu, Nasim, and Mian]{liu2024context}
Weijia Liu, Bo~Miao, Jiuxin Cao, Xuelin Zhu, Jiawei Ge, Bo~Liu, Mehwish Nasim, and Ajmal Mian.
\newblock Context-enhanced video moment retrieval with large language models.
\newblock \emph{IEEE Trans. Multimedia}, pages 1--11, 2025.

\bibitem[Li et~al.(2025{\natexlab{a}})Li, Ding, Cheng, Li, Wang, and Gao]{li2025etc}
Guozhang Li, Xinpeng Ding, De~Cheng, Jie Li, Nannan Wang, and Xinbo Gao.
\newblock Etc: Temporal boundary expand then clarify for weakly supervised video grounding with multimodal large language model.
\newblock \emph{IEEE Trans. Multimedia}, 27:\penalty0 1772--1782, 2025{\natexlab{a}}.

\bibitem[Schuhmann et~al.(2022)Schuhmann, Beaumont, Vencu, Gordon, Wightman, Cherti, Coombes, Katta, Mullis, Wortsman, et~al.]{laion-5b}
Christoph Schuhmann, Romain Beaumont, Richard Vencu, Cade Gordon, Ross Wightman, Mehdi Cherti, Theo Coombes, Aarush Katta, Clayton Mullis, Mitchell Wortsman, et~al.
\newblock Laion-5b: An open large-scale dataset for training next generation image-text models.
\newblock In \emph{NeurIPS}, 2022.

\bibitem[Sharma et~al.(2018)Sharma, Ding, Goodman, and Soricut]{CC}
Piyush Sharma, Nan Ding, Sebastian Goodman, and Radu Soricut.
\newblock Conceptual captions: A cleaned, hypernymed, image alt-text dataset for automatic image captioning.
\newblock In \emph{ACL}, 2018.

\bibitem[Chen et~al.(2024)Chen, Wang, Tian, Ye, Gao, Cui, Tong, Hu, Luo, Ma, et~al.]{internvl}
Zhe Chen, Weiyun Wang, Hao Tian, Shenglong Ye, Zhangwei Gao, Erfei Cui, Wenwen Tong, Kongzhi Hu, Jiapeng Luo, Zheng Ma, et~al.
\newblock How far are we to gpt-4v? closing the gap to commercial multimodal models with open-source suites.
\newblock \emph{Science China Information Sciences}, 2024.

\bibitem[Bai et~al.(2023)Bai, Bai, Yang, Wang, Tan, Wang, Lin, Zhou, and Zhou]{qwenvl}
Jinze Bai, Shuai Bai, Shusheng Yang, Shijie Wang, Sinan Tan, Peng Wang, Junyang Lin, Chang Zhou, and Jingren Zhou.
\newblock Qwen-vl: A versatile vision-language model for understanding, localization, text reading, and beyond.
\newblock \emph{arXiv:2308.12966}, 2023.

\bibitem[Cheng et~al.(2024)Cheng, Leng, Zhang, Xin, Li, Chen, Zhu, Zhang, Luo, Zhao, and Bing]{videollama2}
Zesen Cheng, Sicong Leng, Hang Zhang, Yifei Xin, Xin Li, Guanzheng Chen, Yongxin Zhu, Wenqi Zhang, Ziyang Luo, Deli Zhao, and Lidong Bing.
\newblock Videollama 2: Advancing spatial-temporal modeling and audio understanding in video-llms.
\newblock \emph{arXiv:2406.07476}, 2024.

\bibitem[Li et~al.(2024{\natexlab{b}})Li, Wang, He, Li, Wang, Liu, Wang, Xu, Chen, Luo, et~al.]{mvbench}
Kunchang Li, Yali Wang, Yinan He, Yizhuo Li, Yi~Wang, Yi~Liu, Zun Wang, Jilan Xu, Guo Chen, Ping Luo, et~al.
\newblock Mvbench: A comprehensive multi-modal video understanding benchmark.
\newblock In \emph{CVPR}, 2024{\natexlab{b}}.

\bibitem[Kim et~al.(2024)Kim, Choi, Lee, and Rhee]{ig-vlm}
Wonkyun Kim, Changin Choi, Wonseok Lee, and Wonjong Rhee.
\newblock An image grid can be worth a video: Zero-shot video question answering using a vlm.
\newblock \emph{IEEE Access}, 2024.

\bibitem[Xu et~al.(2024{\natexlab{a}})Xu, Gao, Gan, Chen, Lai, Gang, Kang, and Dehghan]{slowfast-llava}
Mingze Xu, Mingfei Gao, Zhe Gan, Hong-You Chen, Zhengfeng Lai, Haiming Gang, Kai Kang, and Afshin Dehghan.
\newblock Slowfast-llava: A strong training-free baseline for video large language models.
\newblock \emph{arXiv:2407.15841}, 2024{\natexlab{a}}.

\bibitem[Han et~al.(2024)Han, Guo, Tang, He, Wu, and Wang]{free-video-llm}
Kai Han, Jianyuan Guo, Yehui Tang, Wei He, Enhua Wu, and Yunhe Wang.
\newblock Free video-llm: Prompt-guided visual perception for efficient training-free video llms.
\newblock \emph{arXiv:2410.10441}, 2024.

\bibitem[Lin et~al.(2024)Lin, Ye, Zhu, Cui, Ning, Jin, and Yuan]{video-llava}
Bin Lin, Yang Ye, Bin Zhu, Jiaxi Cui, Munan Ning, Peng Jin, and Li~Yuan.
\newblock Video-llava: Learning united visual representation by alignment before projection.
\newblock In \emph{EMNLP}, 2024.

\bibitem[Jin et~al.(2024)Jin, Takanobu, Zhang, Cao, and Yuan]{chatuniv}
Peng Jin, Ryuichi Takanobu, Wancai Zhang, Xiaochun Cao, and Li~Yuan.
\newblock Chat-univi: Unified visual representation empowers large language models with image and video understanding.
\newblock In \emph{CVPR}, 2024.

\bibitem[Liu et~al.(2024{\natexlab{a}})Liu, Li, Wu, and Lee]{llava}
Haotian Liu, Chunyuan Li, Qingyang Wu, and Yong~Jae Lee.
\newblock Visual instruction tuning.
\newblock In \emph{NeurIPS}, 2024{\natexlab{a}}.

\bibitem[Maaz et~al.(2024)Maaz, Rasheed, Khan, and Khan]{videochatgpt}
Muhammad Maaz, Hanoona Rasheed, Salman Khan, and Fahad~Shahbaz Khan.
\newblock Video-chatgpt: Towards detailed video understanding via large vision and language models.
\newblock In \emph{ACL}, 2024.

\bibitem[Huang et~al.(2024)Huang, Wang, Chen, Song, and Zhu]{vtimellm}
Bin Huang, Xin Wang, Hong Chen, Zihan Song, and Wenwu Zhu.
\newblock Vtimellm: Empower llm to grasp video moments.
\newblock In \emph{CVPR}, 2024.

\bibitem[Xu et~al.(2024{\natexlab{b}})Xu, Zhao, Zhou, Lin, Ng, and Feng]{pllava}
Lin Xu, Yilin Zhao, Daquan Zhou, Zhijie Lin, See~Kiong Ng, and Jiashi Feng.
\newblock Pllava: Parameter-free llava extension from images to videos for video dense captioning.
\newblock \emph{arXiv:2404.16994}, 2024{\natexlab{b}}.

\bibitem[Zhang et~al.()Zhang, Li, Liu, Lee, Gui, Fu, Feng, Liu, and Li]{llava-next}
Yuanhan Zhang, Bo~Li, Haotian Liu, Yong~Jae Lee, Liangke Gui, Di~Fu, Jiashi Feng, Ziwei Liu, and Chunyuan Li.
\newblock Llava-next: A strong zero-shot video understanding model.
\newblock \url{https://llava-vl.github.io/blog/2024-04-30-llava-next-video}.

\bibitem[Zhang et~al.(2024)Zhang, Zhang, Li, Zeng, Yang, Zhang, Wang, Tan, Li, and Liu]{longva}
Peiyuan Zhang, Kaichen Zhang, Bo~Li, Guangtao Zeng, Jingkang Yang, Yuanhan Zhang, Ziyue Wang, Haoran Tan, Chunyuan Li, and Ziwei Liu.
\newblock Long context transfer from language to vision.
\newblock \emph{arXiv:2406.16852}, 2024.

\bibitem[Chen et~al.(2025)Chen, Xue, Li, Hu, Zhu, Li, Fang, Tang, Yang, Liu, et~al.]{longvila}
Yukang Chen, Fuzhao Xue, Dacheng Li, Qinghao Hu, Ligeng Zhu, Xiuyu Li, Yunhao Fang, Haotian Tang, Shang Yang, Zhijian Liu, et~al.
\newblock Longvila: Scaling long-context visual language models for long videos.
\newblock In \emph{ICLR}, 2025.

\bibitem[Liu et~al.(2024{\natexlab{b}})Liu, Wang, Ma, Wu, Ma, Wei, Jiao, Wu, and Hu]{kangaroo}
Jiajun Liu, Yibing Wang, Hanghang Ma, Xiaoping Wu, Xiaoqi Ma, Xiaoming Wei, Jianbin Jiao, Enhua Wu, and Jie Hu.
\newblock Kangaroo: A powerful video-language model supporting long-context video input.
\newblock \emph{arXiv:2408.15542}, 2024{\natexlab{b}}.

\bibitem[Li et~al.(2024{\natexlab{c}})Li, Wang, and Jia]{llama-vid}
Yanwei Li, Chengyao Wang, and Jiaya Jia.
\newblock Llama-vid: An image is worth 2 tokens in large language models.
\newblock In \emph{ECCV}, 2024{\natexlab{c}}.

\bibitem[Song et~al.(2024)Song, Chai, Wang, Zhang, Zhou, Wu, Chi, Guo, Ye, Zhang, et~al.]{moviechat}
Enxin Song, Wenhao Chai, Guanhong Wang, Yucheng Zhang, Haoyang Zhou, Feiyang Wu, Haozhe Chi, Xun Guo, Tian Ye, Yanting Zhang, et~al.
\newblock Moviechat: From dense token to sparse memory for long video understanding.
\newblock In \emph{CVPR}, 2024.

\bibitem[Ren et~al.(2024)Ren, Yao, Li, Sun, and Hou]{timechat}
Shuhuai Ren, Linli Yao, Shicheng Li, Xu~Sun, and Lu~Hou.
\newblock Timechat: A time-sensitive multimodal large language model for long video understanding.
\newblock In \emph{CVPR}, 2024.

\bibitem[Liang et~al.(2024)Liang, Li, Bai, Huang, Sun, Wang, He, Cui, Chen, and Zhang]{keyvideollm}
Hao Liang, Jiapeng Li, Tianyi Bai, Xijie Huang, Linzhuang Sun, Zhengren Wang, Conghui He, Bin Cui, Chong Chen, and Wentao Zhang.
\newblock Keyvideollm: Towards large-scale video keyframe selection.
\newblock \emph{arXiv:2407.03104}, 2024.

\bibitem[Luo et~al.(2024)Luo, Zheng, Yang, Li, Lin, Huang, Ji, Chao, Luo, and Ji]{videorag}
Yongdong Luo, Xiawu Zheng, Xiao Yang, Guilin Li, Haojia Lin, Jinfa Huang, Jiayi Ji, Fei Chao, Jiebo Luo, and Rongrong Ji.
\newblock Video-rag: Visually-aligned retrieval-augmented long video comprehension.
\newblock \emph{arXiv:2411.13093}, 2024.

\bibitem[Liang et~al.(2025)Liang, Li, Hu, Yu, Zheng, and Lai]{liang2025rethinking}
Tianming Liang, Linhui Li, Jian-Fang Hu, Xiangyang Yu, Wei-Shi Zheng, and Jianhuang Lai.
\newblock Rethinking temporal context in video-qa: A comprehensive study of single-frame static bias.
\newblock \emph{IEEE Trans. Multimedia}, pages 1--15, 2025.

\bibitem[Xu et~al.(2017)Xu, Zhao, Xiao, Wu, Zhang, He, and Zhuang]{msvd-qa}
Dejing Xu, Zhou Zhao, Jun Xiao, Fei Wu, Hanwang Zhang, Xiangnan He, and Yueting Zhuang.
\newblock Video question answering via gradually refined attention over appearance and motion.
\newblock In \emph{ACM MM}, 2017.

\bibitem[Jang et~al.(2017)Jang, Song, Yu, Kim, and Kim]{tgif-qa}
Yunseok Jang, Yale Song, Youngjae Yu, Youngjin Kim, and Gunhee Kim.
\newblock Tgif-qa: Toward spatio-temporal reasoning in visual question answering.
\newblock In \emph{CVPR}, 2017.

\bibitem[Yu et~al.(2019)Yu, Xu, Yu, Yu, Zhao, Zhuang, and Tao]{activity-net-qa}
Zhou Yu, Dejing Xu, Jun Yu, Ting Yu, Zhou Zhao, Yueting Zhuang, and Dacheng Tao.
\newblock Activitynet-qa: A dataset for understanding complex web videos via question answering.
\newblock In \emph{AAAI}, 2019.

\bibitem[Fu et~al.(2025{\natexlab{a}})Fu, Dai, Luo, Li, Ren, Zhang, Wang, Zhou, Shen, Zhang, Chen, Li, Lin, Zhao, Li, Xu, Zheng, Chen, Ji, and Sun]{videomme}
Chaoyou Fu, Yuhan Dai, Yongdong Luo, Lei Li, Shuhuai Ren, Renrui Zhang, Zihan Wang, Chenyu Zhou, Yunhang Shen, Mengdan Zhang, Peixian Chen, Yanwei Li, Shaohui Lin, Sirui Zhao, Ke~Li, Tong Xu, Xiawu Zheng, Enhong Chen, Rongrong Ji, and Xing Sun.
\newblock Video-mme: The first-ever comprehensive evaluation benchmark of multi-modal llms in video analysis.
\newblock In \emph{CVPR}, 2025{\natexlab{a}}.

\bibitem[Liu et~al.(2024{\natexlab{c}})Liu, Li, Liu, Wang, Ren, Li, Chen, Sun, and Hou]{tempcompass}
Yuanxin Liu, Shicheng Li, Yi~Liu, Yuxiang Wang, Shuhuai Ren, Lei Li, Sishuo Chen, Xu~Sun, and Lu~Hou.
\newblock Tempcompass: Do video llms really understand videos?
\newblock In \emph{ACL (Findings)}, 2024{\natexlab{c}}.

\bibitem[Hong et~al.(2025)Hong, Yan, Cai, Jiang, Hu, and Xie]{hong2025worldsense}
Jack Hong, Shilin Yan, Jiayin Cai, Xiaolong Jiang, Yao Hu, and Weidi Xie.
\newblock Worldsense: Evaluating real-world omnimodal understanding for multimodal llms.
\newblock \emph{arXiv:2502.04326}, 2025.

\bibitem[Li et~al.(2024{\natexlab{d}})Li, Fan, Wong, Kankanhalli, and Yang]{topa}
Wei Li, Hehe Fan, Yongkang Wong, Mohan~S Kankanhalli, and Yi~Yang.
\newblock Topa: Extending large language models for video understanding via text-only pre-alignment.
\newblock In \emph{NeurIPS}, 2024{\natexlab{d}}.

\bibitem[Li et~al.(2025{\natexlab{b}})Li, Liu, Yao, Zhang, An, Wang, Sun, Kong, and Liu]{t3}
Lei Li, Yuanxin Liu, Linli Yao, Peiyuan Zhang, Chenxin An, Lean Wang, Xu~Sun, Lingpeng Kong, and Qi~Liu.
\newblock Temporal reasoning transfer from text to video.
\newblock In \emph{ICLR}, 2025{\natexlab{b}}.

\bibitem[Radford et~al.(2021)Radford, Kim, Hallacy, Ramesh, Goh, Agarwal, Sastry, Askell, Mishkin, Clark, et~al.]{clip}
Alec Radford, Jong~Wook Kim, Chris Hallacy, Aditya Ramesh, Gabriel Goh, Sandhini Agarwal, Girish Sastry, Amanda Askell, Pamela Mishkin, Jack Clark, et~al.
\newblock Learning transferable visual models from natural language supervision.
\newblock In \emph{ICML}, 2021.

\bibitem[Ye et~al.(2023)Ye, Hu, Xu, Ye, Yan, Xu, Li, Tian, Qian, Zhang, et~al.]{ureader}
Jiabo Ye, Anwen Hu, Haiyang Xu, Qinghao Ye, Ming Yan, Guohai Xu, Chenliang Li, Junfeng Tian, Qi~Qian, Ji~Zhang, et~al.
\newblock Ureader: Universal ocr-free visually-situated language understanding with multimodal large language model.
\newblock In \emph{EMNLP (Findings)}, 2023.

\bibitem[Li et~al.(2024{\natexlab{e}})Li, Yang, Liu, Ma, Zhang, Yang, Sun, Liu, and Bai]{monkey}
Zhang Li, Biao Yang, Qiang Liu, Zhiyin Ma, Shuo Zhang, Jingxu Yang, Yabo Sun, Yuliang Liu, and Xiang Bai.
\newblock Monkey: Image resolution and text label are important things for large multi-modal models.
\newblock In \emph{CVPR}, 2024{\natexlab{e}}.

\bibitem[Lin et~al.(2023)Lin, Liu, Zhang, Gao, Qiu, Xiao, Qiu, Lin, Shao, Chen, Han, Huang, Zhang, He, Li, and Qiao]{sphinx}
Ziyi Lin, Chris Liu, Renrui Zhang, Peng Gao, Longtian Qiu, Han Xiao, Han Qiu, Chen Lin, Wenqi Shao, Keqin Chen, Jiaming Han, Siyuan Huang, Yichi Zhang, Xuming He, Hongsheng Li, and Yu~Qiao.
\newblock Sphinx: The joint mixing of weights, tasks, and visual embeddings for multi-modal large language models.
\newblock \emph{arXiv:2311.07575}, 2023.

\bibitem[Share14()]{sharegemini}
Share14.
\newblock Sharegemini: Scaling up video caption data for multimodal large language models.
\newblock \url{https://github.com/Share14/ShareGemini}.

\bibitem[Bain et~al.(2021)Bain, Nagrani, Varol, and Zisserman]{webvid}
Max Bain, Arsha Nagrani, G{\"u}l Varol, and Andrew Zisserman.
\newblock Frozen in time: A joint video and image encoder for end-to-end retrieval.
\newblock In \emph{ICCV}, 2021.

\bibitem[GeminiTeam(2024)]{gemini1dot5}
GeminiTeam.
\newblock Gemini 1.5: Unlocking multimodal understanding across millions of tokens of context.
\newblock \emph{arXiv:2403.05530}, 2024.

\bibitem[Bolya et~al.(2023)Bolya, Fu, Dai, Zhang, Feichtenhofer, and Hoffman]{tome}
Daniel Bolya, Cheng-Yang Fu, Xiaoliang Dai, Peizhao Zhang, Christoph Feichtenhofer, and Judy Hoffman.
\newblock Token merging: Your vit but faster.
\newblock In \emph{ICLR}, 2023.

\bibitem[Caba~Heilbron et~al.(2015)Caba~Heilbron, Escorcia, Ghanem, and Carlos~Niebles]{caba2015activitynet}
Fabian Caba~Heilbron, Victor Escorcia, Bernard Ghanem, and Juan Carlos~Niebles.
\newblock Activitynet: A large-scale video benchmark for human activity understanding.
\newblock In \emph{CVPR}, 2015.

\bibitem[Zhou et~al.(2024{\natexlab{a}})Zhou, Liu, Xu, Iyer, Sun, Mao, Ma, Efrat, Yu, Yu, et~al.]{zhou2024lima}
Chunting Zhou, Pengfei Liu, Puxin Xu, Srinivasan Iyer, Jiao Sun, Yuning Mao, Xuezhe Ma, Avia Efrat, Ping Yu, Lili Yu, et~al.
\newblock Lima: Less is more for alignment.
\newblock In \emph{NeurIPS}, 2024{\natexlab{a}}.

\bibitem[Zeng et~al.(2024)Zeng, Zhang, Zheng, Xia, Wei, Wei, Zhang, Kong, and Song]{zeng2024matters}
Yan Zeng, Hanbo Zhang, Jiani Zheng, Jiangnan Xia, Guoqiang Wei, Yang Wei, Yuchen Zhang, Tao Kong, and Ruihua Song.
\newblock What matters in training a gpt4-style language model with multimodal inputs?
\newblock In \emph{NAACL}, 2024.

\bibitem[Xu et~al.(2025)Xu, Jiang, Niu, Deng, Poovendran, Choi, and Lin]{magpie}
Zhangchen Xu, Fengqing Jiang, Luyao Niu, Yuntian Deng, Radha Poovendran, Yejin Choi, and Bill~Yuchen Lin.
\newblock Magpie: Alignment data synthesis from scratch by prompting aligned llms with nothing.
\newblock In \emph{ICLR}, 2025.

\bibitem[Zhao et~al.(2024)Zhao, Ren, Hessel, Cardie, Choi, and Deng]{zhao2024wildchat}
Wenting Zhao, Xiang Ren, Jack Hessel, Claire Cardie, Yejin Choi, and Yuntian Deng.
\newblock Wildchat: 1m chatgpt interaction logs in the wild.
\newblock In \emph{ICLR}, 2024.

\bibitem[OpenAI({\natexlab{a}})]{gpt-4v}
OpenAI.
\newblock Gpt-4v(ision) system card.
\newblock \url{https://cdn.openai.com/papers/GPTV_System_Card.pdf}, {\natexlab{a}}.

\bibitem[Anthropic()]{claude-3.5-sonnet}
Anthropic.
\newblock Introducing claude 3.5 sonnet.
\newblock \url{https://www.anthropic.com/news/claude-3-5-sonnet}.

\bibitem[OpenAI({\natexlab{b}})]{gpt-4o}
OpenAI.
\newblock Hello gpt-4o.
\newblock \url{https://openai.com/index/hello-gpt-4o}, {\natexlab{b}}.

\bibitem[Team et~al.(2023)Team, Anil, Borgeaud, Alayrac, Yu, Soricut, Schalkwyk, Dai, Hauth, Millican, et~al.]{gemini-1.5-pro}
Gemini Team, Rohan Anil, Sebastian Borgeaud, Jean-Baptiste Alayrac, Jiahui Yu, Radu Soricut, Johan Schalkwyk, Andrew~M Dai, Anja Hauth, Katie Millican, et~al.
\newblock Gemini: a family of highly capable multimodal models.
\newblock \emph{arXiv:2312.11805}, 2023.

\bibitem[Fu et~al.(2024)Fu, Lin, Long, Shen, Zhao, Zhang, Wang, Yin, Ma, Zheng, et~al.]{vita}
Chaoyou Fu, Haojia Lin, Zuwei Long, Yunhang Shen, Meng Zhao, Yifan Zhang, Xiong Wang, Di~Yin, Long Ma, Xiawu Zheng, et~al.
\newblock Vita: Towards open-source interactive omni multimodal llm.
\newblock \emph{arXiv:2408.05211}, 2024.

\bibitem[Fu et~al.(2025{\natexlab{b}})Fu, Lin, Wang, Zhang, Shen, Liu, Li, Long, Gao, Li, et~al.]{fu2025vita}
Chaoyou Fu, Haojia Lin, Xiong Wang, Yi-Fan Zhang, Yunhang Shen, Xiaoyu Liu, Yangze Li, Zuwei Long, Heting Gao, Ke~Li, et~al.
\newblock Vita-1.5: Towards gpt-4o level real-time vision and speech interaction.
\newblock \emph{arXiv:2501.01957}, 2025{\natexlab{b}}.

\bibitem[Wu et~al.(2024)Wu, Li, Chen, and Li]{longvideobench}
Haoning Wu, Dongxu Li, Bei Chen, and Junnan Li.
\newblock Longvideobench: A benchmark for long-context interleaved video-language understanding.
\newblock In \emph{NeurIPS}, 2024.

\bibitem[Zhou et~al.(2024{\natexlab{b}})Zhou, Shu, Zhao, Wu, Xiao, Yang, Xiong, Zhang, Huang, and Liu]{mlvu}
Junjie Zhou, Yan Shu, Bo~Zhao, Boya Wu, Shitao Xiao, Xi~Yang, Yongping Xiong, Bo~Zhang, Tiejun Huang, and Zheng Liu.
\newblock Mlvu: A comprehensive benchmark for multi-task long video understanding.
\newblock \emph{arXiv:2406.04264}, 2024{\natexlab{b}}.

\bibitem[Meta()]{llama}
Meta.
\newblock Introducing llama 3.1: Our most capable models to date.
\newblock \url{https://ai.meta.com/blog/meta-llama-3-1}.

\bibitem[Duan et~al.(2024)Duan, Yang, Qiao, Fang, Chen, Liu, Dong, Zang, Zhang, Wang, et~al.]{vlmevalkit}
Haodong Duan, Junming Yang, Yuxuan Qiao, Xinyu Fang, Lin Chen, Yuan Liu, Xiaoyi Dong, Yuhang Zang, Pan Zhang, Jiaqi Wang, et~al.
\newblock Vlmevalkit: An open-source toolkit for evaluating large multi-modality models.
\newblock In \emph{ACM MM}, 2024.

\end{thebibliography}
